\documentclass{article} 
\usepackage{iclr2025_conference,times}

\usepackage{amsmath,amsfonts,bm}

\def\eqref#1{equation~\ref{#1}}

\def\1{\bm{1}}

\DeclareMathAlphabet{\mathsfit}{\encodingdefault}{\sfdefault}{m}{sl}
\SetMathAlphabet{\mathsfit}{bold}{\encodingdefault}{\sfdefault}{bx}{n}

\usepackage{multirow}
\usepackage{graphicx}
\usepackage[normalem]{ulem}
\useunder{\uline}{\ul}{}
\usepackage{todonotes}
\usepackage[hidelinks]{hyperref}
\usepackage{amsmath}
\usepackage{cleveref}
\usepackage{multicol}
\usepackage{longtable}
\usepackage[labelformat=simple]{subcaption}

\usepackage{wrapfig}
\usepackage{booktabs}
\usepackage{diagbox}
\usepackage{enumitem} 
\usepackage[export]{adjustbox}
\usepackage[misc]{ifsym}
\newcommand{\bench}{GIFT-Eval}
\newcommand{\naive}{\texttt{Naive}}
\newcommand{\snaive}{\texttt{Seasonal Naive}}
\newcommand{\arima}{\texttt{Auto\_Arima}}
\newcommand{\ets}{\texttt{Auto\_ETS}}
\newcommand{\thetaf}{\texttt{Auto\_Theta}}

\newcommand{\deepar}{\texttt{DeepAR}}
\newcommand{\tft}{\texttt{TFT}}
\newcommand{\tide}{\texttt{TiDE}}
\newcommand{\nbeats}{\texttt{N-BEATS}}
\newcommand{\crossformer}{\texttt{Crossformer}}
\newcommand{\patchtst}{\texttt{PatchTST}}
\newcommand{\dlinear}{\texttt{DLinear}}
\newcommand{\itransformer}{\texttt{iTransformer}}

\newcommand{\moirai}{\texttt{Moirai}}

\newcommand{\moiraismall}{\moirai{}\textsubscript{Small}}
\newcommand{\moiraibase}{\moirai{}\textsubscript{Base}}
\newcommand{\moirailarge}{\moirai{}\textsubscript{Large}}

\newcommand{\omoirai}{\texttt{Moirai-Leakage}}

\newcommand{\chronos}{\texttt{Chronos}}
\newcommand{\chronostiny}{\chronos{}\textsubscript{Tiny}}
\newcommand{\chronossmall}{\chronos{}\textsubscript{Small}}
\newcommand{\chronosbase}{\chronos{}\textsubscript{Base}}
\newcommand{\chronoslarge}{\chronos{}\textsubscript{Large}}
\newcommand{\timesfm}{\texttt{TimesFM}}
\newcommand{\visionts}{\texttt{VisionTS}}

\newcommand{\tnaive}{\textbf{\texttt{Nv.}}}
\newcommand{\tsnaive}{\textbf{\texttt{S.Nv.}}}
\newcommand{\tarima}{\textbf{\texttt{A.Ar.}}}
\newcommand{\tets}{\textbf{\texttt{A.ETS}}}
\newcommand{\tthetaf}{\textbf{\texttt{A.Th.}}}

\newcommand{\tdeepar}{\textbf{\texttt{DeepAR}}}
\newcommand{\ttft}{\textbf{\texttt{TFT}}}
\newcommand{\ttide}{\textbf{\texttt{TiDE}}}
\newcommand{\tnbeats}{\textbf{\texttt{N-BEATS}}}
\newcommand{\tcrossformer}{\textbf{\texttt{C.former}}}
\newcommand{\tpatchtst}{\textbf{\texttt{P.TST}}}
\newcommand{\tdlinear}{\textbf{\texttt{DLin.}}}
\newcommand{\titransformer}{\textbf{\texttt{iTrans.}}}

\newcommand{\ttimesfm}{\textbf{\texttt{T.FM}}}
\newcommand{\tvisionts}{\textbf{\texttt{Vis.TS}}}
\newcommand{\tchronos}{\textbf{\texttt{Chr.}}}

\newcommand{\tchronossmall}{\textbf{\tchronos{}\textsubscript{S}}}
\newcommand{\tchronosbase}{\textbf{\tchronos{}\textsubscript{B}}}
\newcommand{\tchronoslarge}{\textbf{\tchronos{}\textsubscript{L}}}

\newcommand{\tmoirai}{\textbf{\texttt{Moi.}}}
\newcommand{\tomoirai}{\textbf{\texttt{Moi Leak.}}}

\newcommand{\tmoiraismall}{\textbf{\tmoirai{}\textsubscript{S}}}
\newcommand{\tmoiraibase}{\textbf{\tmoirai{}\textsubscript{B}}}
\newcommand{\tmoirailarge}{\textbf{\tmoirai{}\textsubscript{L}}}

\newcommand{\tomoiraismall}{\textbf{\tomoirai{}\textsubscript{S}}}
\newcommand{\tomoiraibase}{\textbf{\tomoirai{}\textsubscript{B}}}
\newcommand{\tomoirailarge}{\textbf{\tomoirai{}\textsubscript{L}}}

\newcommand{\heatmapwidth}{0.48\textwidth}

\definecolor{taa-color}{rgb}{0,0,1}
\definecolor{HIGHLIGHT-COLOR}{rgb}{0.9, 0.1, 0.1}

\newcommand{\ta}[1]{}
\newcommand{\tatodo}[1]{}
\newcommand{\myhl}[1]{}

\usepackage[utf8]{inputenc} 
\usepackage[T1]{fontenc}    
\usepackage{hyperref}       
\usepackage{url}            
\usepackage{booktabs}       
\usepackage{amsfonts}       
\usepackage{nicefrac}       
\usepackage{microtype}      
\usepackage{xcolor}         
\usepackage{lipsum}

\title{\bench{}: A Benchmark for General Time Series Forecasting Model Evaluation}

\author{Taha Aksu$^{1}$, Gerald Woo$^{1*}$, Juncheng Liu$^{1}$, Xu Liu$^{1,2}$\thanks{Work done during industrial PhD/internship at Salesforce AI Research.}, \ Chenghao Liu$^{1}$\thanks{Corresponding author. Email: chenghao.liu@salesforce.com},\\
\textbf{Silvio Savarese$^{1}$, Caiming Xiong$^{1}$, Doyen Sahoo$^{1}$} \\
$^{1}$Salesforce AI Research, $^{2}$National University of Singapore \\
\texttt{\{iaksu,juncheng.liu,xu.liu,chenghao.liu\}@salesforce.com} \\
\texttt{\{ssavarese,cxiong,dsahoo\}@salesforce.com} \\
\texttt{woogerald@yahoo.com.sg}
}

\iclrfinalcopy

\begin{document}

\maketitle

\begin{abstract}
Time series foundation models excel in zero-shot forecasting, handling diverse tasks without explicit training. However, the advancement of these models has been hindered by the lack of comprehensive benchmarks. To address this gap, we introduce the \textbf{G}eneral T\textbf{I}me Series \textbf{F}orecas\textbf{T}ing Model \textbf{Eval}uation,~\bench{}, a pioneering benchmark aimed at promoting evaluation across diverse datasets. ~\bench{} encompasses 23 datasets over 144,000 time series and 177 million data points, spanning seven domains, 10 frequencies, multivariate inputs, and prediction lengths ranging from short to long-term forecasts. To facilitate the effective pretraining and evaluation of foundation models, we also provide a non-leaking pretraining dataset containing approximately 230 billion data points. Additionally, we provide a comprehensive analysis of 17 baselines, which includes statistical models, deep learning models, and foundation models. We discuss each model in the context of various benchmark characteristics and offer a qualitative analysis that spans both deep learning and foundation models. We believe the insights from this analysis, along with access to this new standard zero-shot time series forecasting benchmark, will guide future developments in time series foundation models. Code, data, and the leaderboard can be found at~\url{https://github.com/SalesforceAIResearch/gift-eval}.

\end{abstract}

\section{Introduction}
The success of foundation model pretraining in language and vision modalities has catalyzed similar progress in time series forecasting. By pretraining on extensive time series datasets, a universal forecasting model can be developed, equipped to address varied downstream forecasting tasks across multiple domains, frequencies, prediction lengths, and number of variates in a zero-shot manner~\citep{woo2024unifiedtraininguniversaltime,rasul2024lagllamafoundationmodelsprobabilistic,chronos}.

A critical aspect of foundation model research is creating a high-quality benchmark that includes large, diverse evaluation data, and preferably non-leaking pretraining data to fairly evaluate models and identify their weaknesses.
Research in Natural Language Processing (NLP) has produced key benchmarks such as GLUE, MMLU,~\textit{etc.}~\citep{wang2019gluemultitaskbenchmarkanalysis,hendrycks2021measuringmassivemultitasklanguage,srivastava2023imitationgamequantifyingextrapolating,chen2021evaluatinglargelanguagemodels}, which are crucial for developing high-quality models.

Unlike NLP, time series foundation models lack a unified, diverse benchmark for fair comparison. For instance, \citet{woo2024unifiedtraininguniversaltime} introduces LOTSA, which remains the largest collection of time series forecasting pre-training data to date. However, the proposed architecture, \moirai{}, is evaluated on existing benchmarks that are tailored to specific forecasting tasks, such as the LSF~\citep{Zhou2020InformerBE} dataset for long-term forecast, and the Monash~\citep{godahewa2021monash} dataset for univariate forecasts. Both datasets lack sufficient diversity in time series characteristics and forecasting tasks, making it challenging to evaluate the zero-shot capabilities of foundation models in handling broad and generalized forecasting scenarios. This limitation remains in the recent empirical evaluations of other foundation models, including those featured in benchmarks such as TimesFM, Chronos, and Lag-Llama~\citep{das2024decoderonlyfoundationmodeltimeseries, chronos, rasul2024lagllamafoundationmodelsprobabilistic}. Furthermore, the inconsistency in pretraining, training, and test splits across various foundation models complicates comparisons and poses a risk of data leakage during in-domain and out-of-domain evaluations. To accelerate the advancement for research on time series model, it is essential to establish a high-quality and diverse benchmark that supports universal forecasting evaluation.

To fill identified gaps, we introduce the \textbf{G}eneral T\textbf{I}me Series \textbf{F}orecas\textbf{T}ing Model \textbf{Eval}uation (\bench{}), consisting of distinct pretraining and train/test components. The pretraining component features 88 datasets including 240 billion data points (\Cref{app_pretrain_data} lists more details on pretraining data). The train/test component features 23 datasets encompassing 144,000 time series and 177 million data points across seven domains and 10 frequencies, with prediction lengths ranging from short to long-term, as well as univariate and multivariate forecasting settings.
Prior to our work, \citet{qiu2024tfbcomprehensivefairbenchmarking} introduced TFB, a comprehensive dataset for time series forecasting. While it offered diversity in the number of variates and domains, it lacks 
the evaluation of foundation models and accompanying pretraining data without leakage. 
Our benchmark fills these gaps and it also includes a broader range of frequencies, a more diverse taxonomy, and a wider span of prediction lengths. We compare \bench{} with other similar benchmarks in~\Cref{tab:bench_comparison}. Our contributions are three-fold:
\begin{itemize}[leftmargin=*]
    \item \textbf{\bench{}:} We introduce a general time series forecasting benchmark that evaluates the zero-shot and universal forecasting capabilities of foundation models. We provide pretraining and train-test components that ensure diversity across multiple characteristics and time series features.
    \item \textbf{Comprehensive Benchmarking:} We design diverse forecasting tasks and evaluate 17 baselines that encompass statistical, deep learning, and foundational models on \bench{}.
    \item \textbf{Detailed Analysis:} We provide insights into the strengths of different models on all aspects of \bench{} including domains, frequencies, prediction lengths, and the number of variates. We further provide a qualitative analysis showing failure cases of both deep learning and foundation models. We believe these insights will contribute to the future development of foundation models. 
\end{itemize}


\section{Related work}

\begin{table}
\caption{Property comparisons of various forecasting benchmarks.}
\label{tab:bench_comparison}
\resizebox{\textwidth}{!}{%
    \begin{tabular}{l|c|c|c|c|c|c|c}
    \toprule
    \multirow{2}{*}{\diagbox{\textbf{Benchmark}}{\textbf{Property}}} &
    \multicolumn{3}{c|}{Data} &
    \multicolumn{2}{c|}{Forecasting Task} &
    \multicolumn{2}{c}{Evaluation} \\ \cline{2-8}
    & Freq. Range & Num. of Domain & Pretraining data & Num. of var. & Pred. Len. & Benchmark Methods & Prob. Forecasting \\
    \midrule
    Monash~\citep{godahewa2021monash} & Secondly $\sim$ Yearly & 7  & No  & Uni & Short & Stat./DL & No  \\ \hline
    TFB~\citep{qiu2024tfbcomprehensivefairbenchmarking}  & Minutely $\sim$ Yearly & 6  & No  & Uni/Multi & Short & Stat./DL  & No  \\ \hline
    LTSF~\citep{Zeng2022AreTE} &Minutely $\sim$ Weekly  &5  & No  & Multi & Long & Stat./DL & No  \\ \hline
    BasicTS+~\citep{Shao2023ExploringPI}  &Minutely $\sim$ Daily  & 3 & No & Multi & Short/Long & Stat./DL & No \\ \hline 
    GIFT-Eval (our work) & Secondly $\sim$ Yearly & 7 & Yes & Uni/Multi & Short/Long & Stat./DL/FM & Yes \\
    \bottomrule
\end{tabular}%
}
\end{table}

\paragraph{Forecasting Methods} Time series forecasters can be broadly categorized into statistical methods, deep learning methods, and, more recently, foundation models. Statistical models rely solely on historical data statistics to predict future values. Among these, ARIMA~\citep{box1970distribution}, ETS~\citep{hyndman2008forecasting}, Theta~\citep{garza2022statsforecast}, and VAR~\citep{godahewa2021monash} are some of the most widely used ones. With the advent of deep learning technologies, models that apply these techniques to time series forecasting have emerged. Examples include DeepAR~\citep{Flunkert2017DeepARPF}, N-BEATS~\citep{Oreshkin2019NBEATSNB}, and DLinear~\citep{Zeng2022AreTE}, which utilize pre-transformer architectures. Additionally, transformer-based models such as PatchTST~\citep{Nie2022ATS}, Autoformer~\citep{Wu2021AutoformerDT}, and Crossformer~\citep{zhang2023crossformer} have been developed. In the last few years, foundation models have been proposed, inspired by their success in other modalities like language and vision. The multivariate Moirai~\citep{woo2024unifiedtraininguniversaltime} forecaster, for instance, is based on an encoder-decoder architecture pretrained on a large dataset. Conversely, Chronos~\citep{chronos} and TimesFM~\citep{das2024decoderonlyfoundationmodeltimeseries} are univariate forecasters trained using a decoder-only model. VisionTS~\citep{Chen2024VisionTSVM} is another univariate forecaster, unique in that it reformulates the forecasting task as an image reconstruction problem, leveraging a pretrained model from the image modality. However, the main bottleneck in building and evaluating these foundation models is the lack of a diverse and large benchmark dataset.

\paragraph{Forecasting Benchmarks} 
To address this challenge, several efforts have been made to develop extensive time series benchmarks. ~\citet{woo2024unifiedtraininguniversaltime} introduced LOTSA, which holds the title for the largest collection of open time series datasets, encompassing 231 billion data points across nine domains. Despite its vast size, the evaluation datasets reuse existing benchmarks from the time series forecasting community and still lack sufficient variety in terms of time series data characteristics and forecasting tasks, which our benchmark aims to augment. \citet{chronos} developed a dataset specifically structured for pretraining, in-domain evaluation, and zero-shot evaluation splits. However, their work is constrained by a limited range of prediction lengths (from 6 to 56), which excludes long-term forecasts, and it restricts the data to univariate forecasting. In contrast, our benchmark encompasses extensive multivariate scenarios and evaluates diverse data across various domains and frequencies. The corpus by \citet{rasul2024lagllamafoundationmodelsprobabilistic} presents a diverse array of domains, yet it comprises only univariate datasets totaling 8,000 time series. In contrast, \bench{} dramatically expands this scope with 144,000 time series, enhancing the breadth and depth of the dataset. The benchmark by \citet{qiu2024tfbcomprehensivefairbenchmarking} is closely aligned with our work in its aim to curate a diverse and comprehensive set of data. However, it lacks pretraining data, does not evaluate foundation models, and limits the taxonomy to time series features only. Our benchmark not only includes pretraining data (with zero-shot evaluation support) but also provides evaluations for foundation models and offers a taxonomy over both characteristics and time series properties. In summary, our benchmark, \bench{}, builds upon and seeks to address the gaps identified in existing time series forecasting benchmarks. We provide a wider comparison with more benchmarks in~\Cref{tab:bench_comparison}. By providing a more diverse and extensive dataset, we aim to facilitate the development and evaluation of foundation models in time series forecasting.
    
\section{\bench{}}
In this section, we first provide a background on time series forecasting tasks and define key characteristics and features of time series data. We then outline the design decisions behind the development of \bench{}, concluding with an analysis that highlights the key features of its final distribution.

\subsection{Background}
We start by defining univariate and multivariate forecasting tasks. After that, we outline the fundamental characteristics of time series datasets which also influenced our data collection process, including domain, frequency, number of variates, and prediction length. We also introduce time series features as part of our data analysis.

\subsubsection{Time Series Forecasting}
Time series forecasting is a task of predicting future values over one (univariate) or more (multivariate) variates given historical (most commonly real-valued) data which is sampled at regular time intervals. Suppose $D = (Y^i,Z^i)_{i=1}^N$ is a dataset of N time series where $Y^i =(y^i_1,y^i_2,\dots,y^i_{T_i}) \in \mathbb{R}^{d_{y_i} \times T_i}$ is the target time series with $d_{y_i}$ variates and $T_i$ time steps and $Z^i = (z^i_1,z^i_2,\dots,z^i_{T_i}) \in \mathbb{R}^{d_{z_i} \times T_i}$ are the set of covaraites with $d_{z_i}$ variates. Then the forecasting task can be modeled as the predictive distribution: $p(Y_{t:t+h} | Y_{t-l:t},Z_{t-l:t+h})$ where $l$ is the context length, and $h$ is the forecast horizon. Univariate forecasting is a special case where the target series is univariate (\textit{i.e.,} $d_{y_i} = 1 $), no covariates are used (\textit{i.e.,} $Z = \emptyset$), and only the historical values of the target time series are utilized for prediction.
\subsubsection{Time Series Characteristics and Features}
\paragraph{Characteristics} Time series datasets possess inherent characteristics that define their structure, and common patterns observed in the data and even choices of modelling techniques. We believe a universal forecasting model should be able to perform irrespective of the domain from which the data is sourced, the granularity at which it was sampled, the length of the forecast horizon and whether it is univariate or multivariate. Thus in our study, we focus on these four characteristics: $(i)$~\textit{Domain}, i.e., the field or industry from which the time series data originates, such as finance, healthcare or meteorology. The domain often has a direct effect on the nature of patterns. 
Another crucial aspect is  $(ii)$ the ~\textit{frequency} of observations, indicating the time intervals at which the data points are recorded -- such as hourly, daily, monthly or annually. $(iii)$~\textit{Prediction length}, or forecast horizon, is the number of future time steps for which predictions are expected. 
Lastly, $(iv)$ the ~\textit{number of variates} pertains to the dimensionality of the time series data. A \textit{univariate} time series consists of observations of a single variable over time, whereas a \textit{multivariate} time series involves multiple interrelated variables. The number of variates adds complexity to the modelling process, as models need to account for dependencies among multiple time series. By ensuring diversity across these specific characteristics in our benchmark, we aim to encompass a wide array of real-life scenarios.
\paragraph{Features} 
Time series features~\footnote{We use the python implementation of tsfeatures library~\citep{tsfeatures_nixtla} to calculate each feature.} are statistical properties that capture essential characteristics of the data. We have selected six such properties to analyze our benchmark, grouped into three categories based on the aspects they assess, \textit{c.f.}~\Cref{app:tsfeatures} for a detailed explanation and formula of each feature. First, we chose two metrics for assessing the temporal attributes of each time series:$(i)$~\textit{Trend} refers to the progression of the time series, indicating whether the data shows an overall increase, decrease or stability over time, where higher values indicate stronger trends.$(ii)$~\textit{Seasonal strength} measures the extent to which regular, repeating patterns occur at specific intervals, such as daily cycles in energy consumption, or annual peaks in finance. The higher the value the more repeating patterns the data exhibits. Second, to assess the forecastability of the time series, we included two metrics:$(iii)$~\textit{Entropy} measures the “forecastability” of a time series, where low values indicate a high signal-to-noise ratio and high values occur when a series is difficult to forecast. $(iv)$~\textit{Hurst exponent} quantifies the long-term memory or persistence in a time series. It indicates whether future values are likely to be influenced by past trends, revert to the mean, or behave randomly, where higher values indicate more persistence. Lastly, to understand the regularity and variability within the time series, we selected two metrics: $(v)$~\textit{Stability} assesses the inconsistency of the mean of the time series. In simpler terms, it can be defined as the variance of the means. Note that, unlike what the name suggests, lower values indicate more stable data. Finally, $(vi)$~\textit{Lumpiness} quantifies the variability of the variance across different segments of the time series. A high value of lumpiness indicates significant fluctuations in variability, which can be challenging to model due to the inconsistent behavior of the data.

\subsection{Datasets}
\label{sec:train_test_data}

To evaluate and advance universal time series forecasting methods, we have curated a comprehensive collection of datasets. Our compilation spans a wide array of domains with varying frequencies, numbers of variates, and prediction lengths. This diversity is crucial for assessing the generalization capabilities of forecasting models across different types of time series data. In the following sections, we provide detailed descriptions of \bench{} and its unique splits, outlining their sources, and key properties. We also conduct a detailed analysis on the test data to gain a better understanding of the datasets' characteristics and the distribution of time series features.

\paragraph{Train/Test Data} We curated the train/test portion of \bench{} with 15 univariate and eight multivariate datasets, spanning sevem domains and 10 frequencies, totaling 144,000 time series and 177 million data points. We adhere to established prediction lengths for well-known datasets like M4~\citep{Makridakis2018TheMC}. For other datasets, we establish three prediction settings—short, medium, and long—based on frequency and domain, with medium and long settings extending the short-term length by factors of 10 and 15, respectively. To support models without multivariate forecasting, our framework flattens multivariate datasets for broader compatibility. Data is stored in the Arrow format~\citep{richardson2023arrow}, ensuring efficient integration into deep learning pipelines. Our benchmark features 97 unique triplets of dataset, frequency, and length, with aggregated results for each model reported across these configurations. The sources of each dataset used in train/test split can be found in~\Cref{app:bench}

We structure the evaluation component of our benchmark by dedicating the final 10\% of each dataset in train/test portion to testing, with the rest allocated for training. A non-overlapping rolling evaluation method is employed, setting a predetermined number of windows in the test split, each equal to the dataset's prediction length. The final window of the training data serves as validation for tuning deep learning model hyperparameters.

\paragraph{Analysis and statistics over test data}

~\Cref{tab:domain_stats,tab:freq_stats,tab:predlen_stats,tab:var_stats} present detailed statistics on the number of time series and total observations within each characteristic category of the test benchmark. Specifically, these tables break down the data by domain (Table~\ref{tab:domain_stats}), frequency (Table~\ref{tab:freq_stats}), prediction length (Table~\ref{tab:predlen_stats}), and variate count (Table~\ref{tab:var_stats}), offering a quantitative overview of the dataset's composition. For the reader seeking more granular information, detailed dataset statistics are provided in~\Cref{tab:dataset}.

\begin{table}[htb!]
\caption{\bench{} Test data statistics aggregated by prediction length.}
\label{tab:predlen_stats}
\resizebox{\textwidth}{!}{%
\begin{tabular}{c|ccccccccc|cc|cc}
\toprule
\textbf{Pred. Length} &
  6 &
  8 &
  12 &
  13 &
  14 &
  18 &
  30 &
  48 &
  60 &
  480 &
  600 &
  720 &
  900 \\ \midrule
\# Series &
  22,974 &
  24,629 &
  3,443 &
  359 &
  4,227 &
  48,000 &
  34,398 &
  6,194 &
  22 &
  3,874 &
  22 &
  3,874 &
  22 \\
\# Obs &
  845,109 &
  2,525,512 &
  201,042 &
  371,579 &
  10,023,836 &
  11,246,411 &
  1,447,848 &
  131,125,706 &
  194,369 &
  129,375,020 &
  194,369 &
  129,375,020 &
  194,369 \\ \midrule
\end{tabular}%
}
\end{table}

\begin{table}[htb!]
\scriptsize
\caption{\bench{} Test data statistics aggregated by domain.}
\label{tab:domain_stats}
\resizebox{\textwidth}{!}{%
\begin{tabular}{c|cccccccc}
\toprule
\textbf{Domain} & Econ/Fin   & Energy     & Healthcare & Nature    & Sales   & Transport  & Web/CloudOps & Grand Total \\ \midrule
\# Series       & 99,974     & 2,036      & 1,036      & 32,618    & 3,717   & 1,341      & 3,524        & 144,246     \\
\# Obs          & 25,266,415 & 74,119,755 & 129,408    & 3,154,921 & 671,707 & 38,028,955 & 16,610,251   & 157,981,412 \\ \bottomrule
\end{tabular}%
}
\end{table}

\begin{table}[htb!]
\caption{\bench{} Test data statistics aggregated by frequency.}
\label{tab:freq_stats}
\resizebox{\textwidth}{!}{%
\begin{tabular}{c|ccccccccccc}
\toprule
\textbf{Frequency} & 10S     & 10T       & 15T        & 5T         & A       & D          & H          & M          & Q         & W       & Grand Total \\
\midrule
\# Series          & 22      & 138       & 528        & 2,074      & 22,974  & 38,625     & 3,454      & 51,443     & 24,000    & 988     & 144,246     \\
\# Obs             & 194,369 & 7,253,424 & 52,498,336 & 49,105,728 & 845,109 & 11,471,684 & 22,268,218 & 11,447,453 & 2,406,108 & 490,983 & 157,981,412 \\ \bottomrule
\end{tabular}%
}
\end{table}

\begin{table}[htb!]
\caption{\bench{} Test data statistics aggregated by number of variates.}
\label{tab:var_stats}
\centering
\resizebox{0.6\textwidth}{!}{%
\begin{tabular}{c|ccccc}
\toprule
\textbf{\# Variates} & 1           & 2          & 7       & 21     & Grand Total \\
\midrule
\# Series            & 140,711     & 3,522      & 10      & 3      & 144,246     \\
\# Obs               & 141,133,451 & 16,575,619 & 210,488 & 61,854 & 157,981,412 \\ 
\bottomrule
\end{tabular}%
}
\end{table}

\begin{figure}[htb!]
    \centering
    \begin{subfigure}[b]{\heatmapwidth}
        \includegraphics[width=\textwidth,right]{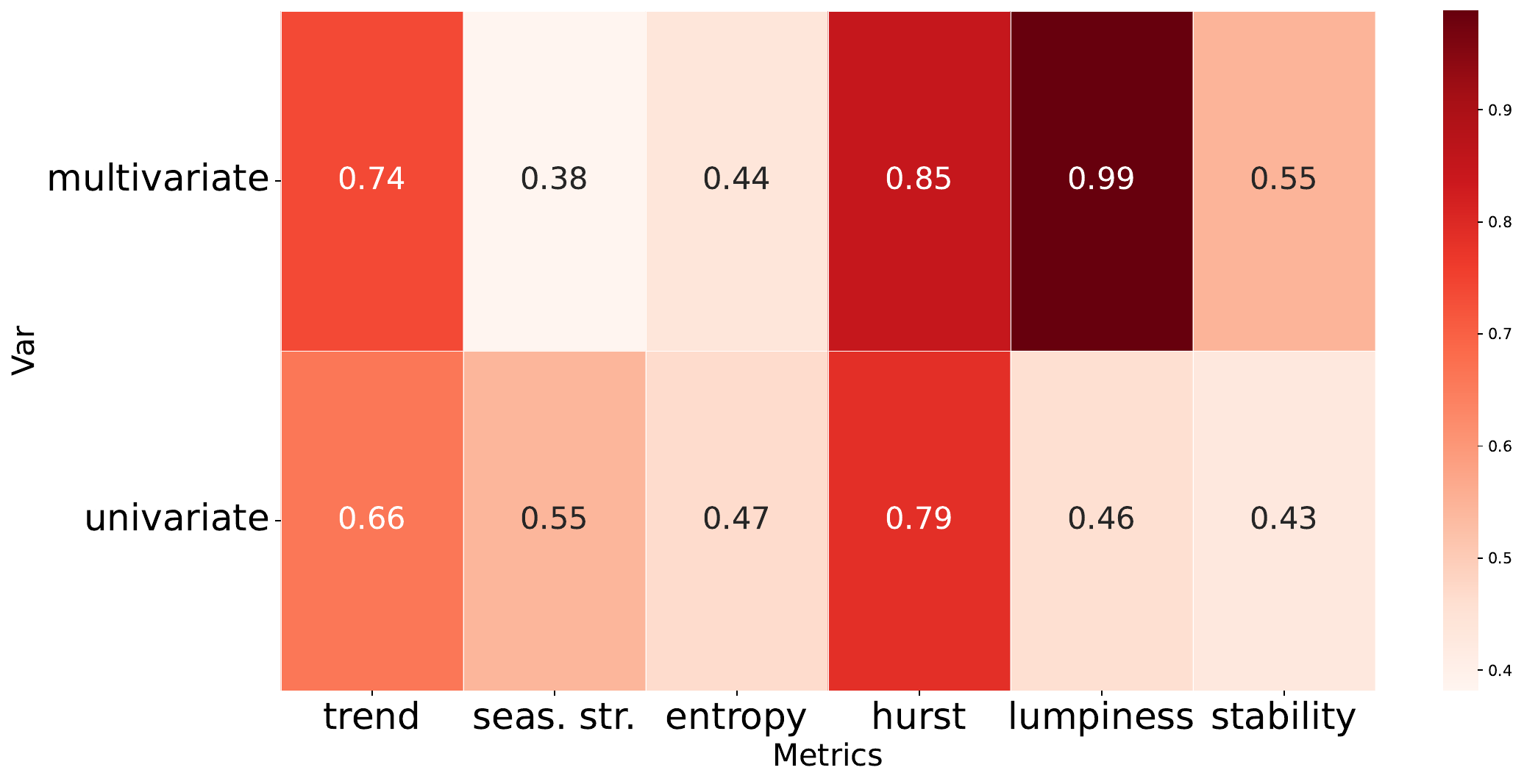}
        \caption{Mean values of TS features across univariate and multivariate datasets.}
        \label{fig:heatmap_var}
    \end{subfigure}%
    \hfill
    \begin{subfigure}[b]{\heatmapwidth}
        \includegraphics[width=\textwidth,left]{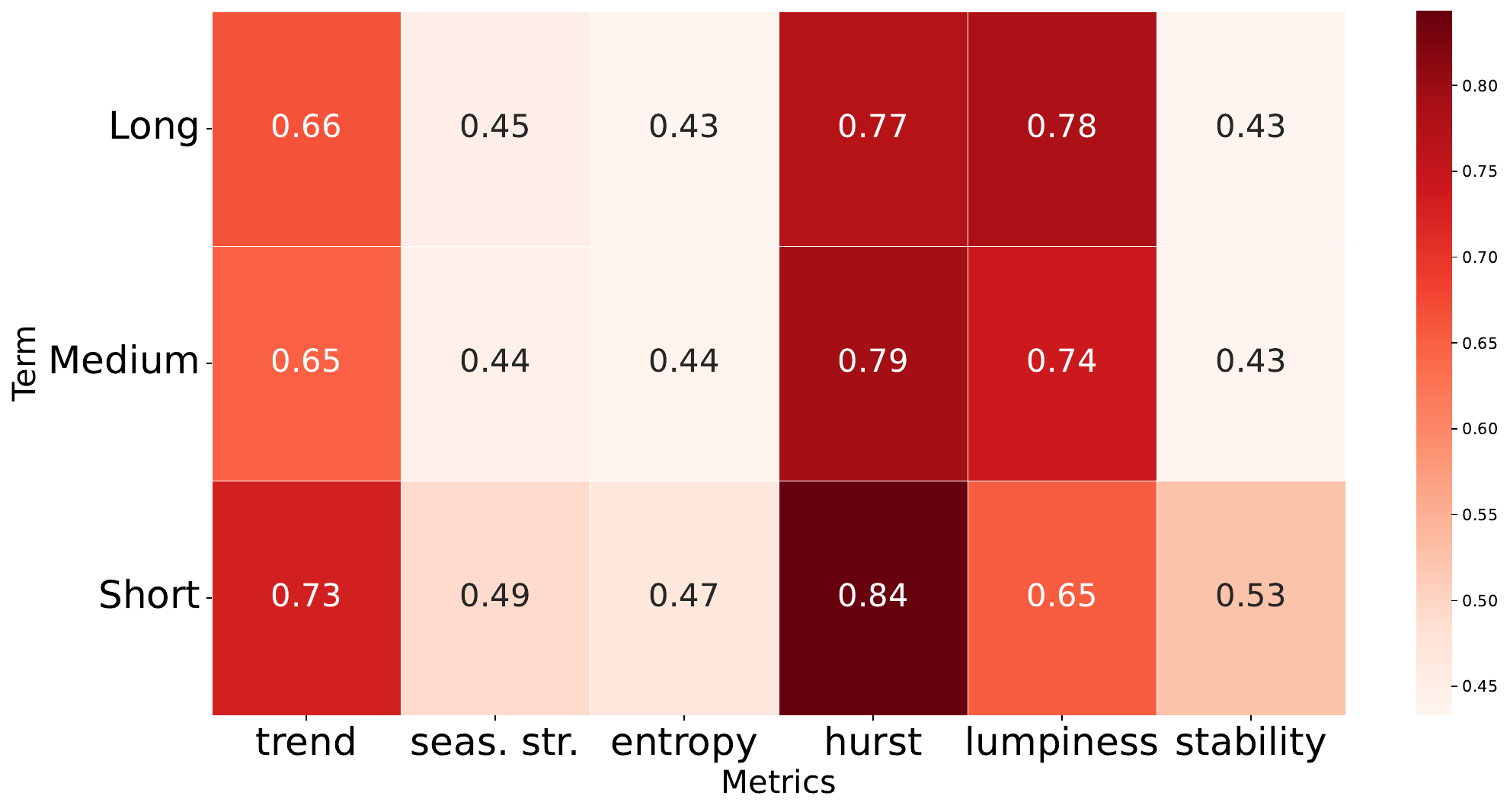}
        \caption{Mean values of TS features across different prediction lengths.}
        \label{fig:heatmap_term}
    \end{subfigure}%
    \hfill
    \begin{subfigure}[b]{\heatmapwidth}
        \includegraphics[width=\textwidth,right]{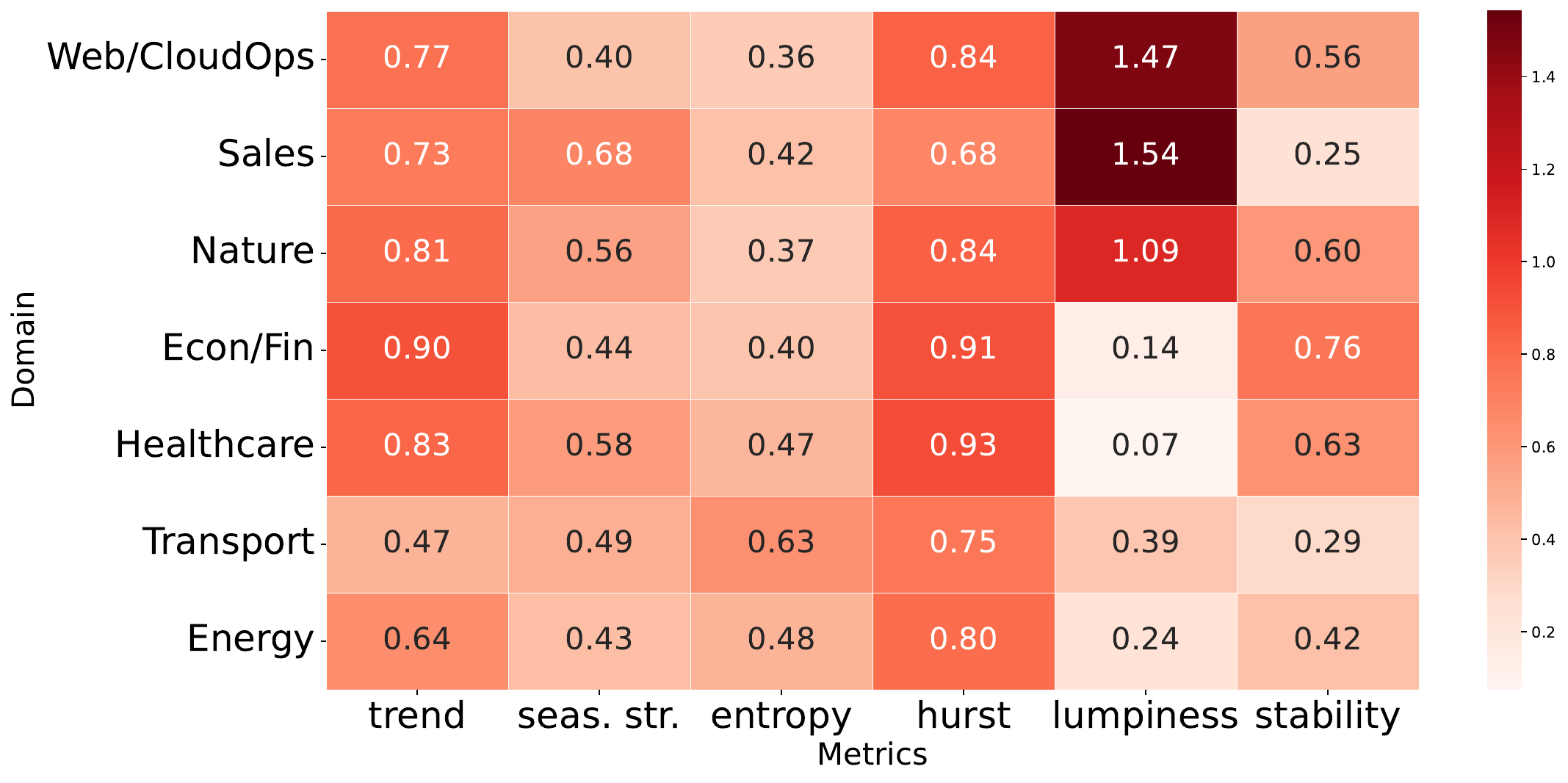}
        \caption{Mean values of TS features across different domains.}
        \label{fig:heatmap_domain}
    \end{subfigure}
    \hfill
    \begin{subfigure}[b]{\heatmapwidth}
        \includegraphics[width=\textwidth,left]{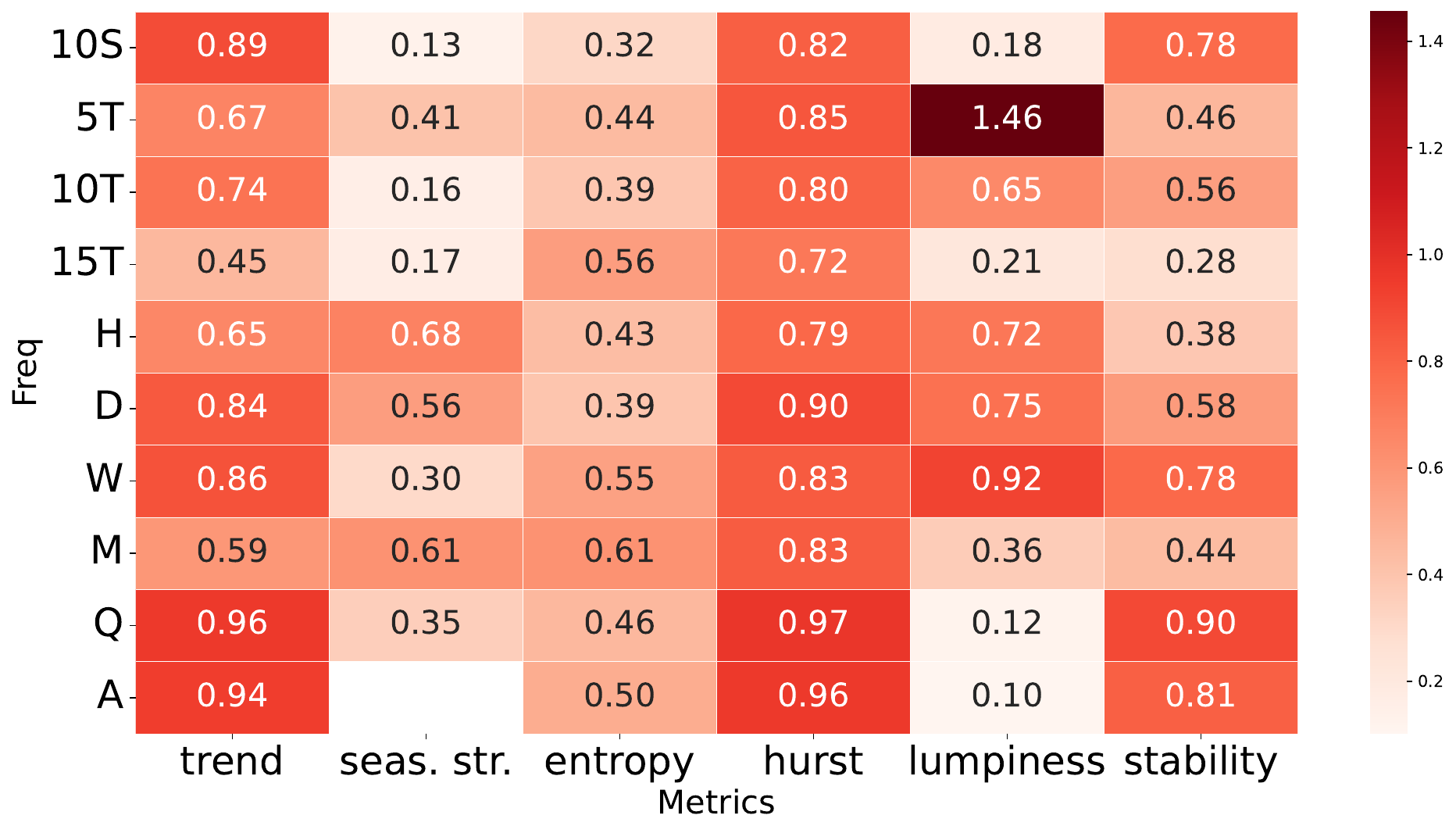}
        \caption{Mean values of TS features across different frequencies.}
        \label{fig:heatmap_freq}
    \end{subfigure}
    \caption{Heatmaps depicting mean values of six time series features across different characteristics.}
    \label{fig:heatmaps}
\end{figure}

We further analyze \bench{} to understand the distribution and characteristics of the time series features across various datasets.~\Cref{fig:heatmaps} illustrates the mean values of each time series features across different dataset characteristics. These heatmaps provide valuable insights into how metrics such as trend, seasonal strength, Hurst exponent, stability, and lumpiness vary across datasets with different domains, frequencies, prediction lengths, and numbers of variates. This visualization aids in identifying patterns and potential biases within the data, ensuring that the benchmark captures a diverse range of time series behaviors. It also facilitates fine-grained analysis of model performance across varying dataset characteristics, offering a comprehensive comparison.

\textbf{Number of variates:} ~\Cref{fig:heatmap_var} depicts that multivariate data exhibit higher stability and lumpiness values, suggesting more fluctuation in variance across different segments, indicating multivariate time series are more complex and potentially more challenging to model. Conversely, univariate series show stronger seasonal strength, reflecting more pronounced and regular repeating patterns, making them more predictable over certain periods. Note that the metrics on multivariate time series are calculated individually for each variate and aggregated for each dataset.

\textbf{Prediction Length:} ~\Cref{fig:heatmap_term} shows that shorter prediction lengths have higher values for both trend and Hurst metrics. This suggests that time series with shorter forecast horizons exhibit stronger directional movements and greater persistence in their trends, making them potentially easier to predict. As the prediction length increases, the trend and hurst values tend to decrease significantly which makes forcasting harder. Notably, the stability values decrease from short to long indicating higher steadiness in long term while lumpiness increases suggesting higher fluctuations in different sections of the data.

\textbf{Domain:} \Cref{fig:heatmap_domain} reveals distinct patterns in the metrics. The Web/CloudOps, Sales, and Nature domains exhibit notably high lumpiness, indicating significant fluctuations in variance. This may reflect the volatile nature of online operations, sales dynamics and weather predictions. On the other hand, Transport shows the highest entropy and lowest trend values, indicating less predictability, likely due to the variable nature of transportation data influenced by numerous external factors. The Econ/Fin domain shows the highest trend values, indicating strong directional movements that may imply clear market trends or economic cycles. Finally, healthcare exhibits the highest Hurst and lowest lumpiness values, suggesting persistence in the data, possibly due to consistent patient trends or medical outcomes over time. 

\textbf{Frequency:} ~\Cref{fig:heatmap_freq} lists frequencties from highest to lowest. Data with very short intervals, such as secondly (S) and minutely (T) exhibit the lowest seasonal strengths and poor steadiness, indicative of the erratic and volatile nature typical at these granular levels. There is a noticeable increase in seasonal strength progressing from secondly and minutely data to hourly (H) and daily (D). Finally, yearly (A) and quarterly (Q) data demonstrate the strongest trends and Hurst values, with notably low lumpiness, suggesting increased persistence and high predictability. Notably, the yearly data lack seasonal strength measurements due to the tsfeatures library not providing seasonal strength for excessively long time series, a limitation commonly observed in low-frequency datasets.

\textbf{Summary:} These observations confirm that our benchmark is rich and diverse, representing a broad range of real-life time series scenarios. Our dataset encompasses various characteristics—such as differing levels of trend, seasonality, persistence, volatility, and complexity—across multiple domains and frequencies. For instance, we include data from domains with high volatility and significant fluctuations, as well as data exhibiting strong persistence and stability. We also cover a wide spectrum of frequencies, from high-frequency data with erratic patterns to low-frequency data with strong trends and greater predictability. In a similar manner, our metrics show diversity across variate types and prediction lengths. This diversity ensures that models are tested across various temporal behaviors, making our benchmark a robust platform for evaluating the general capabilities of unified models, particularly foundation models in time series forecasting.

\paragraph{Pretraining Dataset}
We have also curated a pretraining dataset aligned with \bench{} that has 71 univariate and 17 multivariate datasets, spanning seven domains and 13 frequencies, totaling 4.5 million time series and 230 billion data points. Notably this collection of data has no leakage issue with the train/test split and can be used to pretrain foundation models that can be fairly evaluated on \bench{}. Further details on pretraining dataset can be found in~\Cref{app_pretrain_data}.

\section{Experiments}
In this section, we present the experimental evaluation of \bench{} across various models.

\paragraph{Models}
Time series forecasting training and inference may take different forms for different families of models. Statistical models make predictions by directly analyzing patterns in the historical data without a separate training phase. We incorporate five statistical models in our benchmark: \naive{}, \snaive{}~\citep{hyndman2018forecasting}, \arima{}, \ets{}, and \thetaf{}~\citep{garza2022statsforecast} methods. Deep learning models require training a specific model instance for each dataset. Representing deep learning, we select 8 models: \deepar{}~\citep{Flunkert2017DeepARPF}, \tft{}~\citep{Lim2019TemporalFT}, \tide{}~\citep{Das2023LongtermFW}, \nbeats{}~\citep{Oreshkin2019NBEATSNB}, \patchtst{}~\citep{Nie2022ATS}, \dlinear{}~\citep{Zeng2022AreTE}, \crossformer{}~\citep{zhang2023crossformer} and \itransformer{}~\citep{Liu2023iTransformerIT}.
To obtain both point and probabilistic forecasts, we either adapt models using gluonts~\citep{gluonts_jmlr} with a small probabilistic head or implement our own modifications. We conduct an extensive hyperparameter search for each deep learning model, see~\Cref{app:hyperparam} for details. We evaluate four foundation models on our benchmark:~\timesfm{}~\citep{das2024decoderonlyfoundationmodeltimeseries}, \chronos{}~\citep{chronos} available in tiny, small, and base sizes, ~\moirai{}~\citep{woo2024unifiedtraininguniversaltime} available in small, base, and large sizes and~\visionts{}~\citep{Chen2024VisionTSVM}. These models all provide publicly accessible model parameters for direct use. However, it is important to note that pre-training datasets of ~\timesfm{}, \chronos{}, and ~\moirai{} exhibit partial data leakage issues for \bench{}. Since ~\moirai{} provides pretraining code, here we pretrain a series of ~\moirai{} models using \bench{}'s pretraining split to demonstrate its utility. We empirically investigate the impact of data leakage in Appendix \ref{app:leakage}.
Further details on model-specific hyperparameters and tuning can be found in~\Cref{app:hyperparam}. 

For readability concerns, we omit results from \ets{}, \dlinear{} and \crossformer{} models in the main tables, however, the reader may refer to~\Cref{app:finegrained_results} for results with all models available. For the same space concerns, we use abbreviations to replace each model in the tables. Here is a list of model$\rightarrow$abbreviation pairs for reference: \naive{}: \tnaive{}, \snaive{}: \tsnaive{}, \arima{}: \tarima{}, \thetaf{}: \tthetaf{},\ets{}: \tets{}, \deepar{}: \tdeepar{}, \tft{}: \ttft{}, \tide{}: \ttide{}, \nbeats{}: \tnbeats{}, \patchtst{}: \tpatchtst{}, \itransformer{}: \titransformer{},\dlinear{}: \tdlinear{}, \crossformer{}: \tcrossformer{}, \timesfm{}: \ttimesfm{}, \visionts{}: \tvisionts{}, \chronos{}: \tchronos{}, \chronossmall{}: \tchronossmall{}, \chronosbase{}: \tchronosbase{}, \chronoslarge{}: \tchronoslarge{}, \moirai{}: \tmoirai{}, \moiraismall{}: \tmoiraismall{}, \moiraibase{}: \tmoiraibase{}, \moirailarge{}: \tmoirailarge{}.

\paragraph{Evaluation setting}
Performance is assessed using two metrics: the median Mean Absolute Percentage Error (MAPE) for point forecasts and the Continuous Ranked Probability Score (CRPS)~\citep{gneiting2007strictly} for probabilistic forecasts (definition of both metrics are in~\Cref{app:eval_metrics}). To standardize comparison across benchmarks, both metrics are normalized against the \snaive{} baseline. To avoid skew from any single dataset, we employ a `Rank' metric that assigns a numerical ranking to each model across all 97 configurations judging by their CRPS score. The average of these ranks is then reported as the final Rank for each model.
\subsection{Results}
\label{sec:main_results}
We present results across five distinct parts. The first four parts aggregate the results by the key characteristics that guided the development of our benchmark: domain, prediction length, frequency, and number of variates, then conclude the section with aggregation of results across all configurations. For results on all datasets, frequency and prediction length combinations see~\Cref{tab:all_results,tab:all_results2,tab:all_results3}.
\paragraph{Domain | ~\Cref{tab:domain_results}}

\begin{table}[htb!]
\Large

\caption{Results on \bench{} aggregated by domain. The best results across each row are \textbf{bolded}, while the second best results are \underline{underlined}.}
\label{tab:domain_results}
\resizebox{\textwidth}{!}{%
\begin{tabular}{l|l|llll|llllll|llllllll|l}
\toprule
Domain & Metric & \tnaive{} & \tsnaive{} & \tarima{} & \tthetaf{} & \tdeepar{} & \ttft{} & \ttide{} & \tnbeats{} & \tpatchtst{} & \titransformer{} & \ttimesfm{} & \tvisionts{} & \tchronossmall{} & \tchronosbase{} & \tchronoslarge{} & \tmoiraismall{} & \tmoiraibase{} & \tmoirailarge{} & \textbf{Best} \\
\hline
\multirow{1}{*}{Econ/Fin}& MAPE & 1.22 & 1.00 & $9.37e^{-1}$ & 1.01 & 1.23 & 1.14 & 1.15 & $8.63e^{-1}$ & $9.46e^{-1}$ & 1.02 & $8.41e^{-1}$ & $9.93e^{-1}$ & \underline{$8.02e^{-1}$} & $8.02e^{-1}$ & \boldmath{$7.99e^{-1}$} & 1.04 & 1.27 & $8.80e^{-1}$ & \tchronoslarge{} \\
 & CRPS & 1.36 & 1.00 & $8.25e^{-1}$ & $8.46e^{-1}$ & 1.41 & $8.48e^{-1}$ & 1.09 & $9.86e^{-1}$ & $8.15e^{-1}$ & $8.55e^{-1}$ & \boldmath{$7.29e^{-1}$} & 1.06 & $7.78e^{-1}$ & $7.68e^{-1}$ & $7.74e^{-1}$ & $8.53e^{-1}$ & 1.04 & \underline{$7.36e^{-1}$} & \ttimesfm{} \\
 & Rank & $15.17$ & $14.83$ & $7.17$ & $7.83$ & $14.33$ & $7.83$ & $17.00$ & $12.67$ & $6.67$ & $8.33$ & \underline{$4.83$} & $16.50$ & $7.50$ & $6.33$ & $7.00$ & $10.83$ & $15.50$ & \boldmath{$3.17$} & \tmoirailarge{} \\\hline
\multirow{1}{*}{Energy}& MAPE & 1.23 & 1.00 & 1.02 & 1.35 & 1.68 & 1.09 & 1.30 & 1.17 & 1.00 & 1.24 & $9.90e^{-1}$ & 1.15 & $9.11e^{-1}$ & $9.09e^{-1}$ & $9.12e^{-1}$ & \underline{$8.99e^{-1}$} & $9.87e^{-1}$ & \boldmath{$8.91e^{-1}$} & \tmoirailarge{} \\
 & CRPS & 1.72 & 1.00 & $8.63e^{-1}$ & 2.29 & 1.37 & $6.70e^{-1}$ & $8.82e^{-1}$ & $9.80e^{-1}$ & $6.52e^{-1}$ & $8.26e^{-1}$ & $7.04e^{-1}$ & $8.25e^{-1}$ & $6.85e^{-1}$ & $6.61e^{-1}$ & $6.59e^{-1}$ & $6.97e^{-1}$ & \underline{$6.35e^{-1}$} & \boldmath{$6.26e^{-1}$} & \tmoirailarge{} \\
 & Rank & $18.22$ & $15.84$ & $12.28$ & $16.91$ & $14.69$ & $7.28$ & $10.47$ & $15.09$ & $5.75$ & $7.06$ & $8.25$ & $13.09$ & $8.59$ & $7.38$ & $7.25$ & $7.00$ & \underline{$5.69$} & \boldmath{$5.38$} & \tmoirailarge{} \\\hline
\multirow{1}{*}{Healthcare}& MAPE & 1.21 & 1.00 & $7.83e^{-1}$ & $9.67e^{-1}$ & $8.57e^{-1}$ & $7.90e^{-1}$ & $8.43e^{-1}$ & $7.70e^{-1}$ & $7.95e^{-1}$ & $9.02e^{-1}$ & $7.91e^{-1}$ & $8.03e^{-1}$ & \underline{$6.68e^{-1}$} & $7.25e^{-1}$ & \boldmath{$6.62e^{-1}$} & 1.06 & 1.15 & $8.17e^{-1}$ & \tchronoslarge{} \\
 & CRPS & 1.29 & 1.00 & $6.31e^{-1}$ & $8.28e^{-1}$ & $8.64e^{-1}$ & $7.01e^{-1}$ & 1.16 & $8.69e^{-1}$ & $6.79e^{-1}$ & $7.65e^{-1}$ & $7.99e^{-1}$ & $7.96e^{-1}$ & \underline{$5.75e^{-1}$} & $5.94e^{-1}$ & \boldmath{$5.31e^{-1}$} & $9.26e^{-1}$ & 1.07 & $6.86e^{-1}$ & \tchronoslarge{} \\
 & Rank & $17.80$ & $15.00$ & $7.80$ & $12.00$ & $10.40$ & $8.20$ & $14.40$ & $14.00$ & $8.60$ & $9.40$ & $7.80$ & $12.80$ & $5.80$ & \underline{$5.00$} & \boldmath{$4.20$} & $14.80$ & $15.80$ & $8.00$ & \tchronoslarge{} \\\hline
\multirow{1}{*}{Nature}& MAPE & 1.02 & 1.00 & $9.36e^{-1}$ & 5.15 & 1.34 & 1.35 & 1.65 & 2.05 & $9.76e^{-1}$ & 1.02 & 1.08 & 1.04 & $9.49e^{-1}$ & \underline{$8.09e^{-1}$} & \boldmath{$7.22e^{-1}$} & $8.64e^{-1}$ & 1.08 & $9.01e^{-1}$ & \tchronoslarge{} \\
 & CRPS & 1.52 & 1.00 & $7.10e^{-1}$ & 1.03 & $7.84e^{-1}$ & $4.02e^{-1}$ & $6.49e^{-1}$ & $5.87e^{-1}$ & $4.04e^{-1}$ & $3.92e^{-1}$ & \underline{$3.82e^{-1}$} & $4.87e^{-1}$ & $4.57e^{-1}$ & $4.35e^{-1}$ & $4.33e^{-1}$ & $4.00e^{-1}$ & $4.22e^{-1}$ & \boldmath{$3.78e^{-1}$} & \tmoirailarge{} \\
 & Rank & $19.27$ & $18.60$ & $14.40$ & $16.40$ & $11.87$ & $7.47$ & $13.47$ & $14.53$ & $7.27$ & $6.53$ & $5.73$ & $11.67$ & $9.67$ & $8.73$ & $8.20$ & \underline{$4.40$} & $4.47$ & \boldmath{$4.13$} & \tmoirailarge{} \\\hline
\multirow{1}{*}{Sales}& MAPE & $9.99e^{-1}$ & 1.00 & $7.72e^{-1}$ & $8.26e^{-1}$ & $7.39e^{-1}$ & $7.57e^{-1}$ & 1.00 & $7.26e^{-1}$ & $7.51e^{-1}$ & $7.59e^{-1}$ & $6.83e^{-1}$ & $8.11e^{-1}$ & $7.19e^{-1}$ & $7.01e^{-1}$ & $7.03e^{-1}$ & $6.72e^{-1}$ & \boldmath{$6.68e^{-1}$} & \underline{$6.71e^{-1}$} & \tmoiraibase{} \\
 & CRPS & $9.35e^{-1}$ & 1.00 & $4.83e^{-1}$ & $5.03e^{-1}$ & $3.68e^{-1}$ & \boldmath{$3.64e^{-1}$} & $5.03e^{-1}$ & $4.26e^{-1}$ & $3.67e^{-1}$ & $3.71e^{-1}$ & \underline{$3.65e^{-1}$} & $5.23e^{-1}$ & $3.85e^{-1}$ & $3.85e^{-1}$ & $3.84e^{-1}$ & $4.62e^{-1}$ & $5.16e^{-1}$ & $4.06e^{-1}$ & \ttft{} \\
 & Rank & $19.00$ & $19.25$ & $14.25$ & $14.50$ & $6.25$ & $8.00$ & $14.00$ & $11.00$ & \underline{$3.25$} & $4.75$ & \boldmath{$2.75$} & $15.50$ & $9.00$ & $7.25$ & $7.25$ & $8.50$ & $9.00$ & $5.25$ & \ttimesfm{} \\\hline
\multirow{1}{*}{Transport}& MAPE & 1.48 & 1.00 & 1.07 & 1.08 & $8.11e^{-1}$ & $8.22e^{-1}$ & $8.78e^{-1}$ & \boldmath{$7.60e^{-1}$} & \underline{$8.02e^{-1}$} & $8.27e^{-1}$ & $9.28e^{-1}$ & $8.70e^{-1}$ & $8.46e^{-1}$ & $8.53e^{-1}$ & $8.47e^{-1}$ & $9.10e^{-1}$ & $8.56e^{-1}$ & $9.28e^{-1}$ & \tnbeats{} \\
 & CRPS & 2.29 & 1.00 & $7.90e^{-1}$ & 1.48 & $5.54e^{-1}$ & \underline{$4.85e^{-1}$} & $5.70e^{-1}$ & $6.41e^{-1}$ & $5.06e^{-1}$ & $5.00e^{-1}$ & $5.78e^{-1}$ & $6.66e^{-1}$ & $6.02e^{-1}$ & $5.86e^{-1}$ & $5.85e^{-1}$ & $5.00e^{-1}$ & \boldmath{$4.78e^{-1}$} & $5.02e^{-1}$ & \tmoiraibase{} \\
 & Rank & $20.13$ & $17.33$ & $15.87$ & $18.73$ & $6.47$ & \boldmath{$4.80$} & $10.40$ & $12.60$ & \underline{$5.80$} & $5.87$ & $8.13$ & $13.67$ & $10.07$ & $8.07$ & $8.40$ & $8.33$ & $6.13$ & $6.67$ & \ttft{} \\\hline
\multirow{1}{*}{Web/CloudOps}& MAPE & 1.11 & 1.00 & $8.94e^{-1}$ & $8.33e^{-1}$ & $8.57e^{-1}$ & 1.35 & $9.58e^{-1}$ & \underline{$6.42e^{-1}$} & \boldmath{$6.02e^{-1}$} & $7.19e^{-1}$ & 2.37 & $8.38e^{-1}$ & 1.15 & 1.30 & 1.33 & $7.97e^{-1}$ & 1.06 & $7.91e^{-1}$ & \tpatchtst{} \\
 & CRPS & 1.19 & 1.00 & $9.24e^{-1}$ & $7.38e^{-1}$ & $7.81e^{-1}$ & $6.49e^{-1}$ & $6.73e^{-1}$ & $6.62e^{-1}$ & \boldmath{$5.18e^{-1}$} & \underline{$5.22e^{-1}$} & $9.76e^{-1}$ & $7.24e^{-1}$ & $7.52e^{-1}$ & $8.11e^{-1}$ & $7.91e^{-1}$ & $7.44e^{-1}$ & $7.68e^{-1}$ & $7.42e^{-1}$ & \tpatchtst{} \\
 & Rank & $16.70$ & $16.35$ & $14.50$ & $12.10$ & $11.70$ & $5.95$ & $9.95$ & $10.30$ & \boldmath{$3.95$} & \underline{$4.50$} & $13.90$ & $10.65$ & $10.40$ & $11.35$ & $11.65$ & $8.50$ & $9.25$ & $8.35$ & \tpatchtst{} \\
\bottomrule
\end{tabular}}%

\end{table}

The results across various domains demonstrate that foundation models consistently outperform both statistical and deep learning models. Notably, the \moirai{} and \chronos{} variants achieve top performance in most areas, except in the Web/CloudOps and Transport domains. As discussed in~\Cref{sec:train_test_data} the Transport domain features datasets with the highest entropy and lowest trend components, which are less predictable and Web/CloudOps is one of the domains to exhibit the highest lumpiness. This pattern suggests that foundation models may struggle with time series possessing such characteristics. In contrast, deep learning models like \patchtst{} and \tft{} excel in these challenging domains, possibly indicating a shortfall of the training data used for foundation models in these areas. The comparison of different foundation models yields inconsistent conclusions across various domains. We believe this inconsistency is related to the pre-training datasets utilized by the different foundation models and the varying contribution ratios associated with them.

\paragraph{Prediction length |~\Cref{tab:term_length_results}}

\begin{table}
\Large

\caption{Results on \bench{} aggregated by Prediction Length. The best results across each row are \textbf{bolded}, while the second best results are \underline{underlined}.}
\label{tab:term_length_results}
\resizebox{\textwidth}{!}{%
\begin{tabular}{l|l|llll|llllll|llllllll|l}
\toprule
Pred. Len. & Metric & \tnaive{} & \tsnaive{} & \tarima{} & \tthetaf{} & \tdeepar{} & \ttft{} & \ttide{} & \tnbeats{} & \tpatchtst{} & \titransformer{} & \ttimesfm{} & \tvisionts{} & \tchronossmall{} & \tchronosbase{} & \tchronoslarge{} & \tmoiraismall{} & \tmoiraibase{} & \tmoirailarge{} & \textbf{Best} \\
\hline
\multirow{1}{*}{Long}& MAPE & 1.35 & 1.00 & 1.04 & 3.15 & 1.70 & 1.16 & 1.28 & 1.11 & \underline{$9.62e^{-1}$} & $9.96e^{-1}$ & 1.80 & 1.03 & 1.01 & $9.98e^{-1}$ & 1.00 & \boldmath{$9.19e^{-1}$} & 1.09 & 1.01 & \tmoiraismall{} \\
 & CRPS & 2.11 & 1.00 & $8.43e^{-1}$ & 2.06 & $9.48e^{-1}$ & \underline{$4.53e^{-1}$} & $5.25e^{-1}$ & $6.64e^{-1}$ & \boldmath{$4.37e^{-1}$} & $4.55e^{-1}$ & $6.54e^{-1}$ & $5.46e^{-1}$ & $6.46e^{-1}$ & $6.20e^{-1}$ & $6.21e^{-1}$ & $5.61e^{-1}$ & $5.01e^{-1}$ & $5.39e^{-1}$ & \tpatchtst{} \\
 & Rank & $19.62$ & $16.86$ & $15.10$ & $17.38$ & $12.71$ & \underline{$4.90$} & $8.76$ & $11.95$ & \boldmath{$4.43$} & $5.19$ & $11.29$ & $9.57$ & $11.86$ & $10.71$ & $11.05$ & $7.24$ & $5.43$ & $6.76$ & \tpatchtst{} \\\hline
\multirow{1}{*}{Medium}& MAPE & 1.30 & 1.00 & $9.93e^{-1}$ & 1.87 & 1.07 & 1.26 & 1.19 & 1.15 & \boldmath{$9.03e^{-1}$} & 1.02 & 1.52 & 1.02 & 1.23 & 1.38 & 1.33 & \underline{$9.31e^{-1}$} & 1.11 & $9.60e^{-1}$ & \tpatchtst{} \\
 & CRPS & 2.08 & 1.00 & $8.63e^{-1}$ & 1.96 & $8.27e^{-1}$ & \underline{$5.29e^{-1}$} & $6.11e^{-1}$ & $7.60e^{-1}$ & \boldmath{$5.19e^{-1}$} & $5.29e^{-1}$ & $8.14e^{-1}$ & $6.69e^{-1}$ & $7.38e^{-1}$ & $7.77e^{-1}$ & $7.51e^{-1}$ & $6.61e^{-1}$ & $6.37e^{-1}$ & $6.25e^{-1}$ & \tpatchtst{} \\
 & Rank & $19.29$ & $16.14$ & $14.57$ & $17.43$ & $10.52$ & $4.81$ & $9.52$ & $12.95$ & \boldmath{$4.00$} & \underline{$4.62$} & $10.81$ & $10.95$ & $11.48$ & $11.57$ & $10.90$ & $7.52$ & $6.38$ & $6.76$ & \tpatchtst{} \\\hline
\multirow{1}{*}{Short}& MAPE & 1.10 & 1.00 & $9.18e^{-1}$ & 1.13 & 1.08 & 1.05 & 1.12 & 1.03 & $8.04e^{-1}$ & $9.68e^{-1}$ & $9.29e^{-1}$ & $9.47e^{-1}$ & $7.81e^{-1}$ & \boldmath{$7.50e^{-1}$} & \underline{$7.52e^{-1}$} & $8.49e^{-1}$ & $9.41e^{-1}$ & $7.73e^{-1}$ & \tchronosbase{} \\
 & CRPS & 1.20 & 1.00 & $7.78e^{-1}$ & $9.56e^{-1}$ & 1.03 & $6.76e^{-1}$ & $9.17e^{-1}$ & $8.20e^{-1}$ & $6.28e^{-1}$ & $7.28e^{-1}$ & $6.45e^{-1}$ & $8.20e^{-1}$ & $5.97e^{-1}$ & \underline{$5.90e^{-1}$} & \boldmath{$5.86e^{-1}$} & $6.66e^{-1}$ & $7.08e^{-1}$ & $6.02e^{-1}$ & \tchronoslarge{} \\
 & Rank & $17.22$ & $16.75$ & $11.87$ & $13.53$ & $11.89$ & $8.22$ & $13.42$ & $13.87$ & $6.87$ & $7.49$ & $6.64$ & $14.65$ & $7.25$ & \underline{$6.16$} & $6.27$ & $8.15$ & $8.84$ & \boldmath{$5.40$} & \tmoirailarge{} \\
\bottomrule
\end{tabular}}%
\end{table}

\begin{table}
\Large

\caption{Results on \bench{} aggregated by frequency. The best results across each row are \textbf{bolded}, while second best results are \underline{underlined}.}
\label{tab:frequency_results}
\resizebox{\textwidth}{!}{%
\begin{tabular}{l|l|llll|llllll|llllllll|l}
\toprule
Freq. & Metric & \tnaive{} & \tsnaive{} & \tarima{} & \tthetaf{} & \tdeepar{} & \ttft{} & \ttide{} & \tnbeats{} & \tpatchtst{} & \titransformer{} & \ttimesfm{} & \tvisionts{} & \tchronossmall{} & \tchronosbase{} & \tchronoslarge{} & \tmoiraismall{} & \tmoiraibase{} & \tmoirailarge{} & Best \\
\midrule
10S & CRPS & 1.54 & 1.00 & 1.00 & \boldmath{$4.50e^{-1}$} & 1.03 & 1.05 & $9.85e^{-1}$ & $8.21e^{-1}$ & $7.32e^{-1}$ & \underline{$6.91e^{-1}$} & 1.82 & $9.34e^{-1}$ & 1.10 & 1.27 & 1.19 & 1.31 & 1.40 & 1.24 & \tthetaf{} \\
 & MAPE & 1.97 & 1.00 & 1.00 & \boldmath{$6.49e^{-1}$} & 1.83 & 3.24 & 1.67 & \underline{$9.97e^{-1}$} & 1.06 & 1.31 & 6.23 & 1.10 & 2.42 & 2.95 & 3.02 & 1.65 & 2.34 & 1.55 & \tthetaf{} \\
 & Rank & $16.33$ & $9.50$ & $8.50$ & \boldmath{$1.00$} & $11.17$ & $7.67$ & $10.17$ & $6.67$ & $5.00$ & \underline{$2.50$} & $19.33$ & $10.17$ & $10.83$ & $12.33$ & $11.67$ & $14.83$ & $16.50$ & $14.17$ & \tthetaf{} \\
 \hline
5T & CRPS & 1.40 & 1.00 & 1.00 & 1.05 & $7.97e^{-1}$ & $5.57e^{-1}$ & $6.48e^{-1}$ & $7.26e^{-1}$ & $5.43e^{-1}$ & $5.43e^{-1}$ & $7.30e^{-1}$ & $7.34e^{-1}$ & $7.52e^{-1}$ & $7.49e^{-1}$ & $7.55e^{-1}$ & \underline{$5.36e^{-1}$} & $5.41e^{-1}$ & \boldmath{$5.32e^{-1}$} & \tmoirailarge{} \\
 & MAPE & $8.44e^{-1}$ & 1.00 & 1.00 & $9.64e^{-1}$ & $6.72e^{-1}$ & $6.43e^{-1}$ & $7.95e^{-1}$ & $6.34e^{-1}$ & $5.87e^{-1}$ & $6.60e^{-1}$ & $8.20e^{-1}$ & $8.76e^{-1}$ & $7.80e^{-1}$ & $7.63e^{-1}$ & $7.72e^{-1}$ & \boldmath{$5.30e^{-1}$} & $6.00e^{-1}$ & \underline{$5.52e^{-1}$} & \tmoiraismall{} \\
 & Rank & $17.17$ & $17.75$ & $16.25$ & $16.67$ & $13.25$ & $4.92$ & $10.08$ & $12.42$ & $4.75$ & $5.50$ & $11.92$ & $12.17$ & $11.17$ & $11.25$ & $11.92$ & \underline{$4.42$} & $4.75$ & \boldmath{$3.25$} & \tmoirailarge{} \\
 \hline
10T & CRPS & 2.23 & 1.00 & 1.00 & 3.67 & $5.92e^{-1}$ & \boldmath{$4.12e^{-1}$} & $6.55e^{-1}$ & $7.81e^{-1}$ & $5.58e^{-1}$ & $5.91e^{-1}$ & $5.70e^{-1}$ & $5.01e^{-1}$ & $6.43e^{-1}$ & $5.55e^{-1}$ & $5.51e^{-1}$ & $5.92e^{-1}$ & \underline{$4.33e^{-1}$} & $4.95e^{-1}$ & \ttft{} \\
 & MAPE & $7.59e^{-1}$ & 1.00 & 1.00 & 2.63 & \underline{$6.45e^{-1}$} & 1.20 & 1.16 & 1.26 & $8.81e^{-1}$ & $7.80e^{-1}$ & 1.02 & 1.08 & 1.51 & 1.27 & 1.03 & \boldmath{$5.83e^{-1}$} & $9.12e^{-1}$ & 1.05 & \tmoiraismall{} \\
 & Rank & $19.33$ & $16.50$ & $15.50$ & $19.67$ & $10.67$ & \boldmath{$4.33$} & $11.83$ & $13.17$ & $7.33$ & $6.83$ & $8.33$ & $7.50$ & $12.17$ & $8.83$ & $8.00$ & $8.50$ & \underline{$5.00$} & $7.50$ & \ttft{} \\
 \hline
15T & CRPS & 2.42 & 1.00 & $9.58e^{-1}$ & 1.76 & 1.68 & $7.32e^{-1}$ & $8.07e^{-1}$ & $9.89e^{-1}$ & \underline{$6.63e^{-1}$} & \boldmath{$6.58e^{-1}$} & $7.93e^{-1}$ & $8.78e^{-1}$ & $7.93e^{-1}$ & $7.67e^{-1}$ & $7.64e^{-1}$ & $7.93e^{-1}$ & $7.07e^{-1}$ & $6.76e^{-1}$ & \titransformer{} \\
 & MAPE & 1.44 & 1.00 & 1.10 & $9.89e^{-1}$ & 1.96 & 1.06 & $9.79e^{-1}$ & $9.64e^{-1}$ & \boldmath{$8.74e^{-1}$} & \underline{$9.00e^{-1}$} & 1.02 & $9.82e^{-1}$ & $9.61e^{-1}$ & $9.52e^{-1}$ & $9.44e^{-1}$ & 1.10 & 1.01 & $9.40e^{-1}$ & \tpatchtst{} \\
 & Rank & $19.75$ & $15.17$ & $13.92$ & $17.00$ & $14.50$ & $6.58$ & $10.08$ & $15.00$ & \underline{$3.83$} & \boldmath{$3.50$} & $8.25$ & $12.83$ & $9.75$ & $8.50$ & $8.17$ & $8.67$ & $5.58$ & $4.42$ & \titransformer{} \\
   \hline
H & CRPS & 1.84 & 1.00 & $7.77e^{-1}$ & 1.85 & $8.93e^{-1}$ & $4.87e^{-1}$ & $5.70e^{-1}$ & $6.65e^{-1}$ & \boldmath{$4.62e^{-1}$} & $4.76e^{-1}$ & $5.10e^{-1}$ & $6.11e^{-1}$ & $5.14e^{-1}$ & $5.06e^{-1}$ & $5.09e^{-1}$ & $4.78e^{-1}$ & \underline{$4.63e^{-1}$} & $4.99e^{-1}$ & \tpatchtst{} \\
 & MAPE & 1.33 & 1.00 & $9.54e^{-1}$ & 3.08 & 1.18 & 1.11 & 1.25 & 1.34 & $9.28e^{-1}$ & $9.60e^{-1}$ & $9.82e^{-1}$ & $9.89e^{-1}$ & \boldmath{$7.33e^{-1}$} & \underline{$7.39e^{-1}$} & $7.48e^{-1}$ & $8.13e^{-1}$ & $9.07e^{-1}$ & $8.51e^{-1}$ & \tchronossmall{} \\
 & Rank & $19.65$ & $17.65$ & $15.48$ & $18.87$ & $10.87$ & $6.65$ & $10.58$ & $13.42$ & \boldmath{$4.97$} & $6.13$ & $8.35$ & $11.94$ & $8.58$ & $7.81$ & $8.00$ & $6.32$ & \underline{$5.29$} & $6.55$ & \tpatchtst{} \\
 \hline
D & CRPS & $7.98e^{-1}$ & 1.00 & $4.99e^{-1}$ & $5.75e^{-1}$ & $5.94e^{-1}$ & $4.14e^{-1}$ & $7.48e^{-1}$ & $5.71e^{-1}$ & $4.37e^{-1}$ & $4.95e^{-1}$ & $4.92e^{-1}$ & $5.56e^{-1}$ & $4.37e^{-1}$ & $4.21e^{-1}$ & $4.21e^{-1}$ & \underline{$4.13e^{-1}$} & $4.63e^{-1}$ & \boldmath{$3.92e^{-1}$} & \tmoirailarge{} \\
 & MAPE & 1.00 & 1.00 & $8.53e^{-1}$ & $9.29e^{-1}$ & $8.01e^{-1}$ & $7.62e^{-1}$ & $9.97e^{-1}$ & $7.96e^{-1}$ & $7.52e^{-1}$ & $8.18e^{-1}$ & $7.80e^{-1}$ & $8.95e^{-1}$ & $7.37e^{-1}$ & \underline{$6.86e^{-1}$} & $6.96e^{-1}$ & $7.09e^{-1}$ & $7.45e^{-1}$ & \boldmath{$6.63e^{-1}$} & \tmoirailarge{} \\
 & Rank & $17.60$ & $19.00$ & $11.00$ & $14.33$ & $11.07$ & $6.67$ & $13.53$ & $14.40$ & $7.20$ & $8.47$ & \underline{$5.60$} & $14.80$ & $8.07$ & $6.73$ & $6.67$ & $6.20$ & $7.47$ & \boldmath{$4.33$} & \tmoirailarge{} \\
 \hline
W & CRPS & $8.76e^{-1}$ & 1.00 & $7.49e^{-1}$ & $8.09e^{-1}$ & 1.35 & $7.79e^{-1}$ & 1.22 & 1.08 & $6.98e^{-1}$ & 1.33 & $6.31e^{-1}$ & $9.99e^{-1}$ & \underline{$5.55e^{-1}$} & $5.62e^{-1}$ & \boldmath{$5.52e^{-1}$} & $7.58e^{-1}$ & $7.48e^{-1}$ & $6.41e^{-1}$ & \tchronoslarge{} \\
 & MAPE & 1.00 & 1.00 & $9.76e^{-1}$ & 1.08 & 1.86 & 1.04 & 1.77 & 1.45 & $9.07e^{-1}$ & 1.90 & $8.62e^{-1}$ & 1.14 & \underline{$7.08e^{-1}$} & $7.29e^{-1}$ & \boldmath{$7.07e^{-1}$} & $9.81e^{-1}$ & $9.50e^{-1}$ & $8.96e^{-1}$ & \tchronoslarge{} \\
 & Rank & $13.62$ & $16.50$ & $9.88$ & $12.12$ & $12.75$ & $11.25$ & $13.12$ & $15.25$ & $8.00$ & $12.38$ & $4.88$ & $16.00$ & $5.25$ & \underline{$4.50$} & \boldmath{$4.38$} & $8.62$ & $9.75$ & $5.12$ & \tchronoslarge{} \\
 \hline
M & CRPS & 1.58 & 1.00 & \underline{$7.66e^{-1}$} & $8.82e^{-1}$ & 1.10 & $8.75e^{-1}$ & 1.24 & 1.01 & $8.48e^{-1}$ & $8.21e^{-1}$ & \boldmath{$7.38e^{-1}$} & 1.04 & $8.31e^{-1}$ & $8.61e^{-1}$ & $8.15e^{-1}$ & 1.22 & 1.50 & $9.17e^{-1}$ & \ttimesfm{} \\
 & MAPE & 1.30 & 1.00 & \boldmath{$7.90e^{-1}$} & $9.20e^{-1}$ & 1.18 & 1.05 & 1.04 & $9.41e^{-1}$ & $9.90e^{-1}$ & 1.01 & $8.63e^{-1}$ & $9.38e^{-1}$ & $8.62e^{-1}$ & $8.95e^{-1}$ & \underline{$8.47e^{-1}$} & 1.38 & 1.89 & $9.97e^{-1}$ & \tarima{} \\
 & Rank & $19.20$ & $13.40$ & $6.20$ & $8.40$ & $11.80$ & $8.20$ & $15.40$ & $11.40$ & $7.60$ & \underline{$5.60$} & \boldmath{$3.60$} & $14.80$ & $8.60$ & $8.60$ & $8.20$ & $17.00$ & $19.20$ & $9.60$ & \ttimesfm{} \\
 \hline
Q & CRPS & $9.51e^{-1}$ & 1.00 & $8.23e^{-1}$ & \underline{$7.97e^{-1}$} & $8.41e^{-1}$ & $8.37e^{-1}$ & 1.02 & $9.72e^{-1}$ & $8.35e^{-1}$ & $7.97e^{-1}$ & $8.53e^{-1}$ & 1.05 & $8.46e^{-1}$ & $8.40e^{-1}$ & $8.40e^{-1}$ & $9.32e^{-1}$ & 1.13 & \boldmath{$7.88e^{-1}$} & \tmoirailarge{} \\
 & MAPE & $9.30e^{-1}$ & 1.00 & $8.59e^{-1}$ & $8.38e^{-1}$ & $9.44e^{-1}$ & $9.37e^{-1}$ & 1.13 & $8.38e^{-1}$ & $9.37e^{-1}$ & $9.08e^{-1}$ & $8.80e^{-1}$ & $9.37e^{-1}$ & $8.24e^{-1}$ & \boldmath{$8.10e^{-1}$} & \underline{$8.10e^{-1}$} & 1.04 & 1.43 & $8.87e^{-1}$ & \tchronosbase{} \\
 & Rank & $14.00$ & $16.00$ & $5.00$ & \underline{$2.00$} & $10.00$ & $7.00$ & $17.00$ & $15.00$ & $6.00$ & $3.00$ & $12.00$ & $18.00$ & $11.00$ & $8.00$ & $9.00$ & $13.00$ & $20.00$ & \boldmath{$1.00$} & \tmoirailarge{} \\
 \hline
A & CRPS & $9.93e^{-1}$ & 1.00 & $9.42e^{-1}$ & $8.33e^{-1}$ & $8.19e^{-1}$ & \underline{$7.97e^{-1}$} & 1.12 & $9.71e^{-1}$ & $8.48e^{-1}$ & $8.48e^{-1}$ & $8.48e^{-1}$ & 1.15 & 1.01 & $9.78e^{-1}$ & $9.78e^{-1}$ & $8.26e^{-1}$ & $9.42e^{-1}$ & \boldmath{$7.75e^{-1}$} & \tmoirailarge{} \\
 & MAPE & 1.00 & 1.00 & 1.02 & $9.37e^{-1}$ & 1.02 & \underline{$9.26e^{-1}$} & 1.21 & \boldmath{$9.03e^{-1}$} & 1.01 & 1.03 & $9.71e^{-1}$ & 1.09 & 1.00 & $9.77e^{-1}$ & $9.77e^{-1}$ & $9.77e^{-1}$ & 1.21 & $9.54e^{-1}$ & \tnbeats{} \\
 & Rank & $15.00$ & $16.00$ & $10.00$ & $6.00$ & $4.00$ & \underline{$2.00$} & $18.00$ & $12.00$ & $8.00$ & $7.00$ & $9.00$ & $19.00$ & $17.00$ & $13.00$ & $14.00$ & $5.00$ & $11.00$ & \boldmath{$1.00$} & \tmoirailarge{} \\
\bottomrule
\end{tabular}}%

\end{table}

The analysis by prediction length reveals marked differences in performance across various forecast horizons. For short-term forecasts, foundation models, particularly the \moirai{} variants, consistently outperform other models, underscoring their robustness in handling immediate trends and fluctuations. As the prediction length extends to medium and long terms, the performance of \timesfm{} and \chronos{} have significantly declined. This is because the decoder-only architecture adopts recursive multi-step forecast strategy, leading to severe accumulation error issue. Meanwhile, deep learning models such as \patchtst{} demonstrate superior performance on medium and long terms forecast. This trend indicates that the fine-tuning of foundation models effectively captures longer-term dependencies, which are crucial for accurate predictions over extended periods. Thus despite the progress in foundational time series research, there remains a notable performance gap between deep learning and foundation models for medium to long-term predictions, suggesting an area ripe for further research.

\paragraph{Frequency |~\Cref{tab:frequency_results}}

In the analysis of model performance by data frequency, the highest frequency (secondly) is predominantly led by a statistical baseline, \thetaf{} model, indicating its strength in rapid, granular trend capture. As the frequency decreases, deep learning models like \itransformer{}, \patchtst{}, and \tft{} begin to assert dominance, particularly for minutely and hourly data granularities. However, from daily to yearly frequencies, foundation models consistently outperform other approaches, securing the best results across these settings. This pattern suggests that foundation models, with their extensive pretraining, are better equipped to leverage the broader patterns and slower dynamics typical of lower frequency data, enhancing their forecasting accuracy. Conversely, higher frequency data, which often contains more noise and less discernible patterns, poses challenges that foundation models are less suited to address, as evidenced by their performance relative to more specialized deep learning and statistical models.

\paragraph{Number of variates |~\Cref{tab:variate_results}}

\begin{table}
\Large

\caption{Results on \bench{} aggregated by number of variates. The best results across each row are \textbf{bolded}, while the second best results are \underline{underlined}.}
\label{tab:variate_results}
\resizebox{\textwidth}{!}{%
\begin{tabular}{l|l|llll|llllll|llllllll|l}
\toprule
Num. Var. & Metric & \tnaive{} & \tsnaive{} & \tarima{} & \tthetaf{} & \tdeepar{} & \ttft{} & \ttide{} & \tnbeats{} & \tpatchtst{} & \titransformer{} & \ttimesfm{} & \tvisionts{} & \tchronossmall{} & \tchronosbase{} & \tchronoslarge{} & \tmoiraismall{} & \tmoiraibase{} & \tmoirailarge{} & \textbf{Best} \\
\hline
\multirow{1}{*}{Multi.v}& MAPE & 1.21 & 1.00 & $9.80e^{-1}$ & 2.41 & 1.54 & 1.40 & 1.44 & 1.43 & \boldmath{$8.75e^{-1}$} & 1.15 & 1.71 & 1.04 & 1.08 & 1.11 & 1.10 & \underline{$8.84e^{-1}$} & 1.09 & $9.32e^{-1}$ & \tpatchtst{} \\
 & CRPS & 1.40 & 1.00 & $8.71e^{-1}$ & 1.19 & 1.18 & \underline{$6.09e^{-1}$} & $8.20e^{-1}$ & $7.35e^{-1}$ & \boldmath{$5.27e^{-1}$} & $6.22e^{-1}$ & $7.50e^{-1}$ & $7.01e^{-1}$ & $6.65e^{-1}$ & $6.83e^{-1}$ & $6.73e^{-1}$ & $6.26e^{-1}$ & $6.28e^{-1}$ & $6.27e^{-1}$ & \tpatchtst{} \\
 & Rank & $17.70$ & $16.51$ & $13.91$ & $14.98$ & $14.00$ & $6.91$ & $11.40$ & $12.58$ & \boldmath{$5.02$} & \underline{$5.19$} & $10.07$ & $11.53$ & $9.51$ & $9.44$ & $9.56$ & $6.33$ & $6.28$ & $6.67$ & \tpatchtst{} \\\hline
\multirow{1}{*}{Uni.v}& MAPE & 1.19 & 1.00 & $9.45e^{-1}$ & 1.18 & $9.49e^{-1}$ & $8.92e^{-1}$ & $9.60e^{-1}$ & \boldmath{$7.96e^{-1}$} & $8.47e^{-1}$ & $8.53e^{-1}$ & $8.72e^{-1}$ & $9.34e^{-1}$ & $8.11e^{-1}$ & $8.06e^{-1}$ & \underline{$7.99e^{-1}$} & $8.80e^{-1}$ & $9.47e^{-1}$ & $8.10e^{-1}$ & \tnbeats{} \\
 & CRPS & 1.74 & 1.00 & $7.62e^{-1}$ & 1.59 & $8.02e^{-1}$ & \underline{$5.85e^{-1}$} & $7.23e^{-1}$ & $8.03e^{-1}$ & $5.92e^{-1}$ & $6.28e^{-1}$ & $6.30e^{-1}$ & $7.49e^{-1}$ & $6.18e^{-1}$ & $6.00e^{-1}$ & $5.95e^{-1}$ & $6.55e^{-1}$ & $6.63e^{-1}$ & \boldmath{$5.66e^{-1}$} & \tmoirailarge{} \\
 & Rank & $18.57$ & $16.74$ & $12.56$ & $15.39$ & $10.00$ & $6.65$ & $11.70$ & $13.80$ & \underline{$6.28$} & $7.31$ & $7.33$ & $13.72$ & $8.89$ & $7.43$ & $7.31$ & $9.00$ & $8.59$ & \boldmath{$5.44$} & \tmoirailarge{} \\
\bottomrule
\end{tabular}}%

\end{table}

In multivariate settings, deep learning models, particularly \patchtst{} (best) and \itransformer{} (second best), stand out by delivering the best scores across all evaluated metrics. \moirai{} outperforms other foundation models, as it is the only model that supports multivariate forecasting.
On the other hand, in univariate scenarios, foundation models, especially the large variant of \moirai{}, demonstrate superior performance over their deep learning counterparts. This suggests that foundation models, with their broader pretraining on diverse data sets, are particularly adept at extracting and leveraging predictive signals from single streams of data. 

\paragraph{General |~\Cref{tab:results}}

\begin{table}
\Large

\caption{Results on \bench{} aggregated by all results. The best results across each row are \textbf{bolded}, while the second best results are \underline{underlined}.}
\label{tab:results}
\resizebox{\textwidth}{!}{%
\begin{tabular}{l|llll|llllll|llllllll|l}
\toprule
Metric & \tnaive{} & \tsnaive{} & \tarima{} & \tthetaf{} & \tdeepar{} & \ttft{} & \ttide{} & \tnbeats{} & \tpatchtst{} & \titransformer{} & \ttimesfm{} & \tvisionts{} & \tchronossmall{} & \tchronosbase{} & \tchronoslarge{} & \tmoiraismall{} & \tmoiraibase{} & \tmoirailarge{} & \textbf{Best} \\
\midrule
MAPE & 1.20 & 1.00 & $9.61e^{-1}$ & 1.73 & 1.21 & 1.12 & 1.17 & 1.08 & \boldmath{$8.60e^{-1}$} & $9.85e^{-1}$ & 1.25 & $9.82e^{-1}$ & $9.28e^{-1}$ & $9.40e^{-1}$ & $9.30e^{-1}$ & $8.82e^{-1}$ & 1.01 & \underline{$8.64e^{-1}$} & \tpatchtst{} \\
CRPS & 1.59 & 1.00 & $8.11e^{-1}$ & 1.41 & $9.69e^{-1}$ & $5.96e^{-1}$ & $7.66e^{-1}$ & $7.73e^{-1}$ & \boldmath{$5.63e^{-1}$} & $6.26e^{-1}$ & $6.83e^{-1}$ & $7.28e^{-1}$ & $6.38e^{-1}$ & $6.37e^{-1}$ & $6.29e^{-1}$ & $6.42e^{-1}$ & $6.48e^{-1}$ & \underline{$5.93e^{-1}$} & \tpatchtst{} \\
Rank & $18.19$ & $16.64$ & $13.15$ & $15.21$ & $11.77$ & $6.76$ & $11.57$ & $13.26$ & \boldmath{$5.72$} & $6.37$ & $8.55$ & $12.75$ & $9.16$ & $8.32$ & $8.31$ & $7.81$ & $7.57$ & \underline{$5.99$} & \tpatchtst{} \\
\bottomrule
\end{tabular}}%

\end{table}

\begin{table}
    \Large
    \centering
    \caption{Best and second best counts for each model across~\bench {} dataset configurations (97) according to the Rank metric. The best results across each row are \textbf{bolded}.}
    \label{tab:best_counts}
\resizebox{\textwidth}{!}{
\begin{tabular}{lccccccccccccccccccccc}
\toprule
 & \tmoirailarge{} & \tpatchtst{} & \titransformer{} & \tmoiraibase{} & \tcrossformer{} & \ttimesfm{} & \ttft{} & \tmoiraismall{} & \tchronosbase{} & \tchronoslarge{} & \tthetaf{} & \tdeepar{} & \tarima{} & \tchronossmall{} & \tnbeats{} & \tets{} & \tnaive{} & \tsnaive{} & \ttide{} & \tdlinear{} & \tvisionts{} \\
\midrule
Best & 12 & 10 & 7 & 8 & \textbf{16} & 10 & 11 & 5 & 2 & 6 & 6 & 2 & 2 & 0 & 0 & 0 & 0 & 0 & 0 & 0 & 0 \\
Second Best & 12 & 10 & \textbf{13} & 11 & 2 & 6 & 5 & 10 & 9 & 3 & 1 & 5 & 3 & 3 & 2 & 2 & 0 & 0 & 0 & 0 & 0 \\
Total & \textbf{24} & 20 & 20 & 19 & 18 & 16 & 16 & 15 & 11 & 9 & 7 & 7 & 5 & 3 & 2 & 2 & 0 & 0 & 0 & 0 & 0 \\
\bottomrule
\end{tabular}
}
\end{table}

The comprehensive aggregation of results across the entire benchmark offers insightful performance distinctions. \patchtst{} emerges as the most dominant model, securing the top average scores in all metrics, with \moirailarge{} consistently following in second place. We also present the number of times each model achieves the best or second best results in~\Cref{tab:best_counts}. \crossformer{} appears most frequently as the best performer, and \moirailarge{} as the model that appears in top 2 most frequently. The discrepancy between ranking and count based evaluation suggests that certain datasets may disproportionately influence the metric-based results, which is not captured by the ranking-based outcomes. 
Thus \patchtst{} offers reliable results across diverse datasets, making it a strong generalist. In contrast, \moirailarge{} delivers better performance on particular cases. 

Some recent works ~\citep{Shi2024ScalingLF,Shi2024TimeMoEBT,chronos} have verified
the scaling law in time series foundation models (\textit{i.e.}, larger model performs better), however, \bench{} does not consistently support this conclusion. It only holds in the energy domain or univariate forecasting when only considering ranking metrics.

\subsection{Qualitative Results / Failure Cases}
In addition to the quantitative results discussed earlier, we present qualitative analyses by sharing forecasting samples across various datasets using both deep learning and foundation models. For the deep learning models, we selected four representatives: \patchtst{} and \itransformer{}, from recent transformer-based architectures, and \deepar{} and \nbeats{}, which are more traditional deep learning approaches. Regarding foundational models, we included \moirai{} to represent encoder-decoder architectures, \chronos{} as a decoder-only model, and \visionts{} due to its unique method of representing the time series through image modality. By examining how these models perform on different datasets, we aim to provide deeper insights into their forecasting behaviors, strengths, and limitations.

\begin{figure}[htb!]
    \centering
    \begin{subfigure}[b]{\heatmapwidth}
        \includegraphics[width=\textwidth,right]{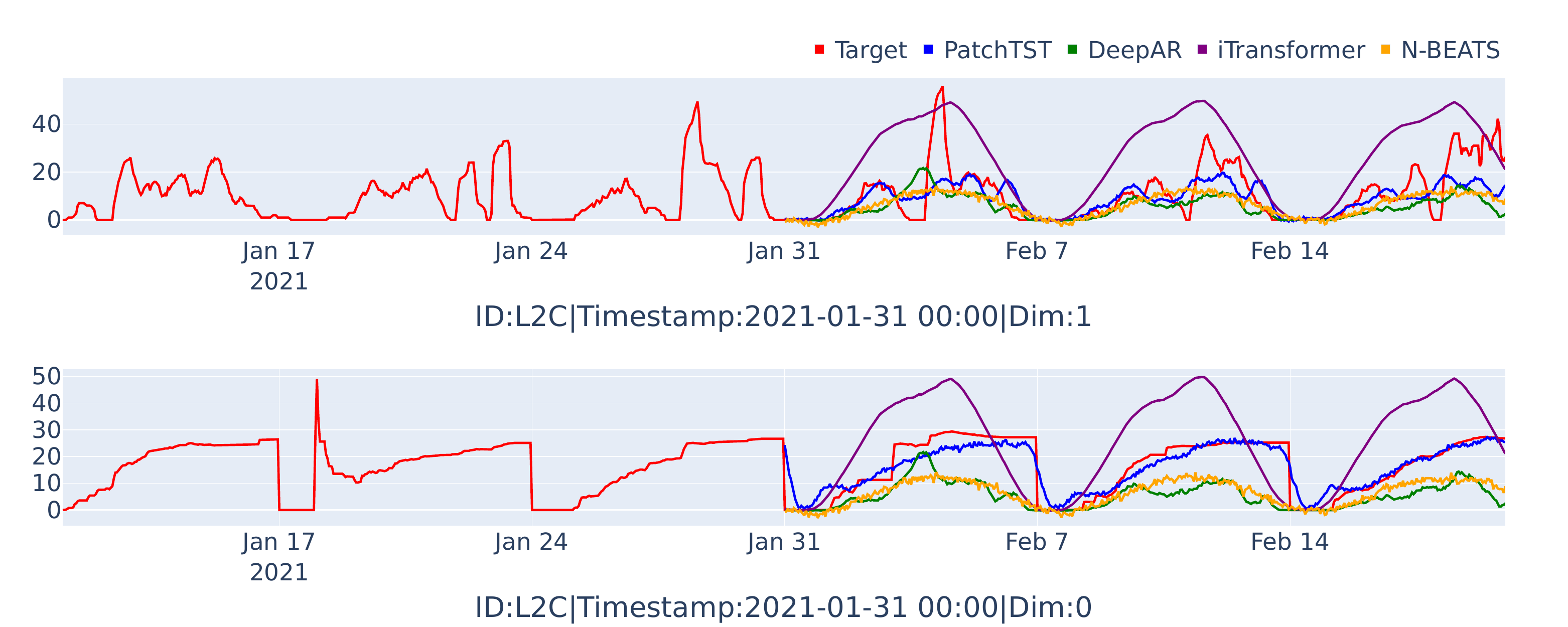}
        \caption{Deep learning forecasts sampled on \textit{Bizitobs\_l2c}, hourly dataset with medium prediction length.}
        \label{fig:qual1}
    \end{subfigure}%
    \hfill
    \begin{subfigure}[b]{\heatmapwidth}
        \includegraphics[width=\textwidth,right]{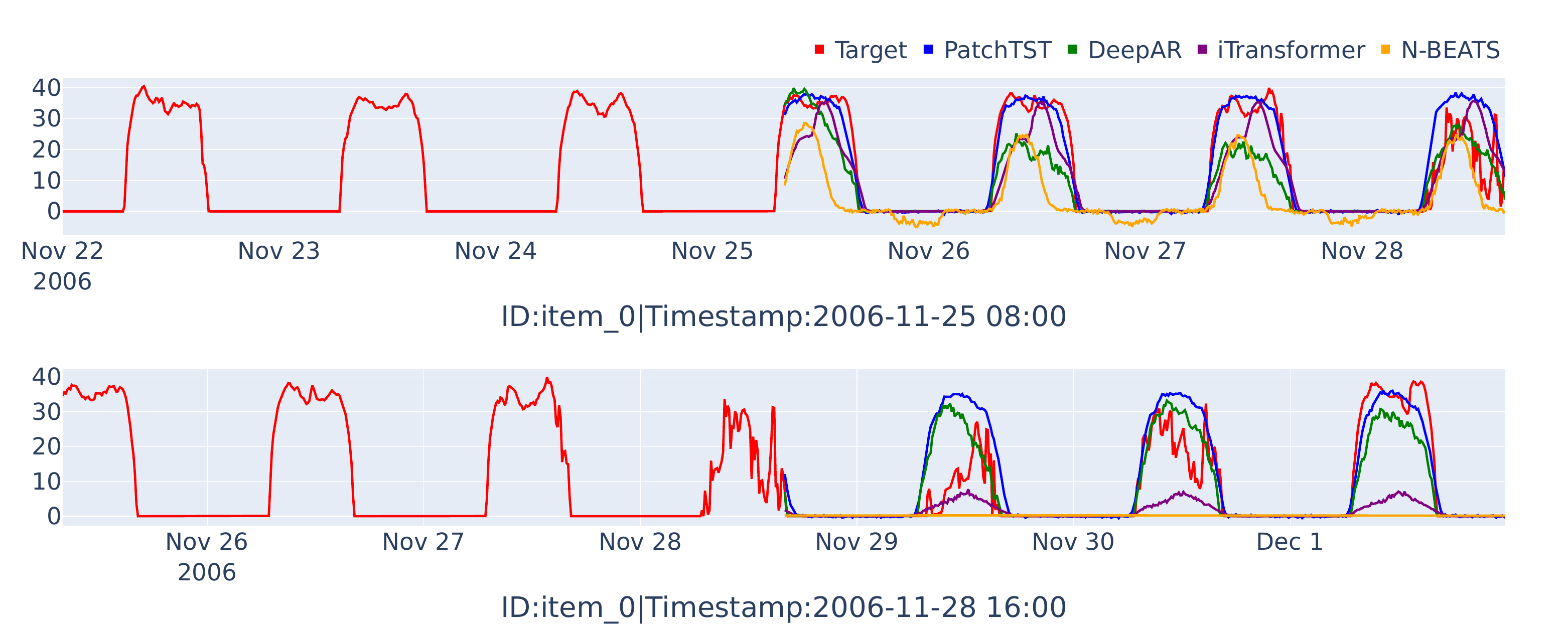}
        \caption{Deep learning forecasts sampled on \textit{Solar} ten--minutely dataset with medium prediction length.}
        \label{fig:qual2}
    \end{subfigure}%
    \hfill
    \begin{subfigure}[b]{\heatmapwidth}
        \includegraphics[width=\textwidth,left]{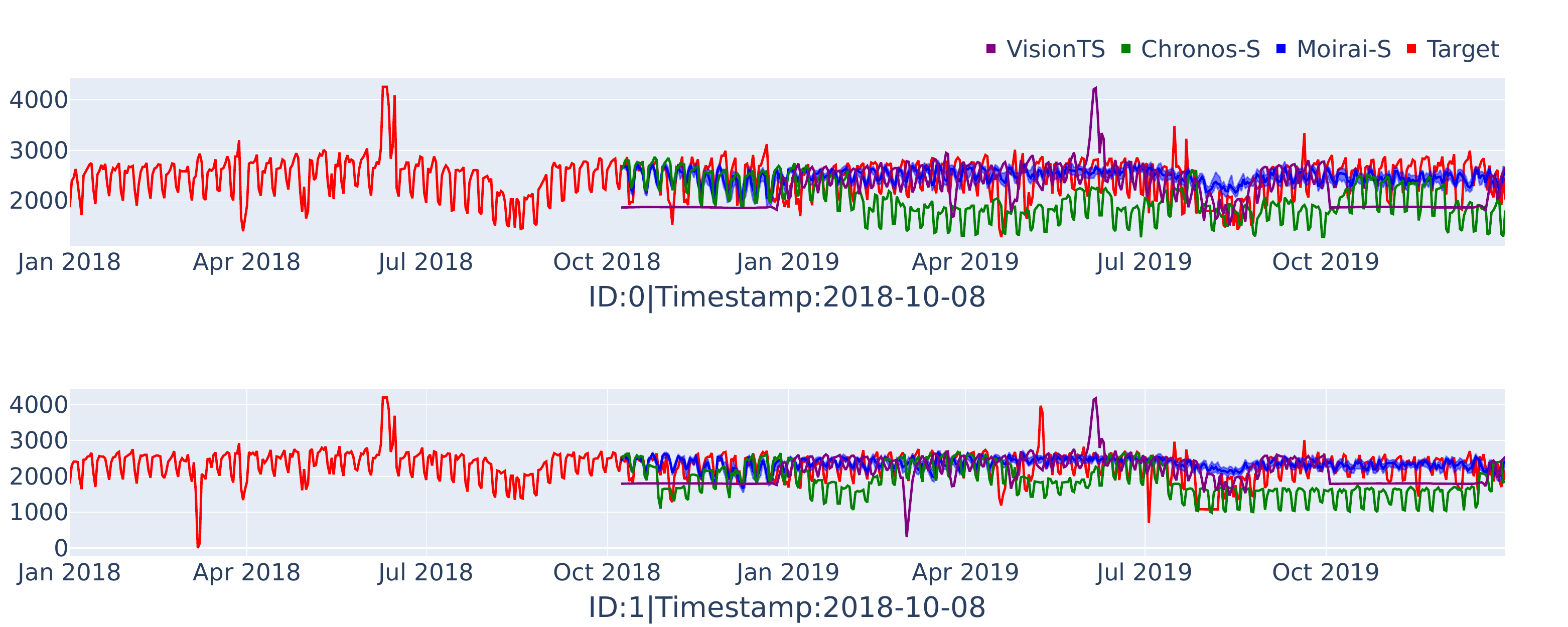}
        \caption{Foundation model forecasts sampled on \textit{M\_DENSE} daily dataset with long prediction length.}
        \label{fig:qual3}
    \end{subfigure}
    \hfill
    \begin{subfigure}[b]{\heatmapwidth}
        \includegraphics[width=\textwidth,left]{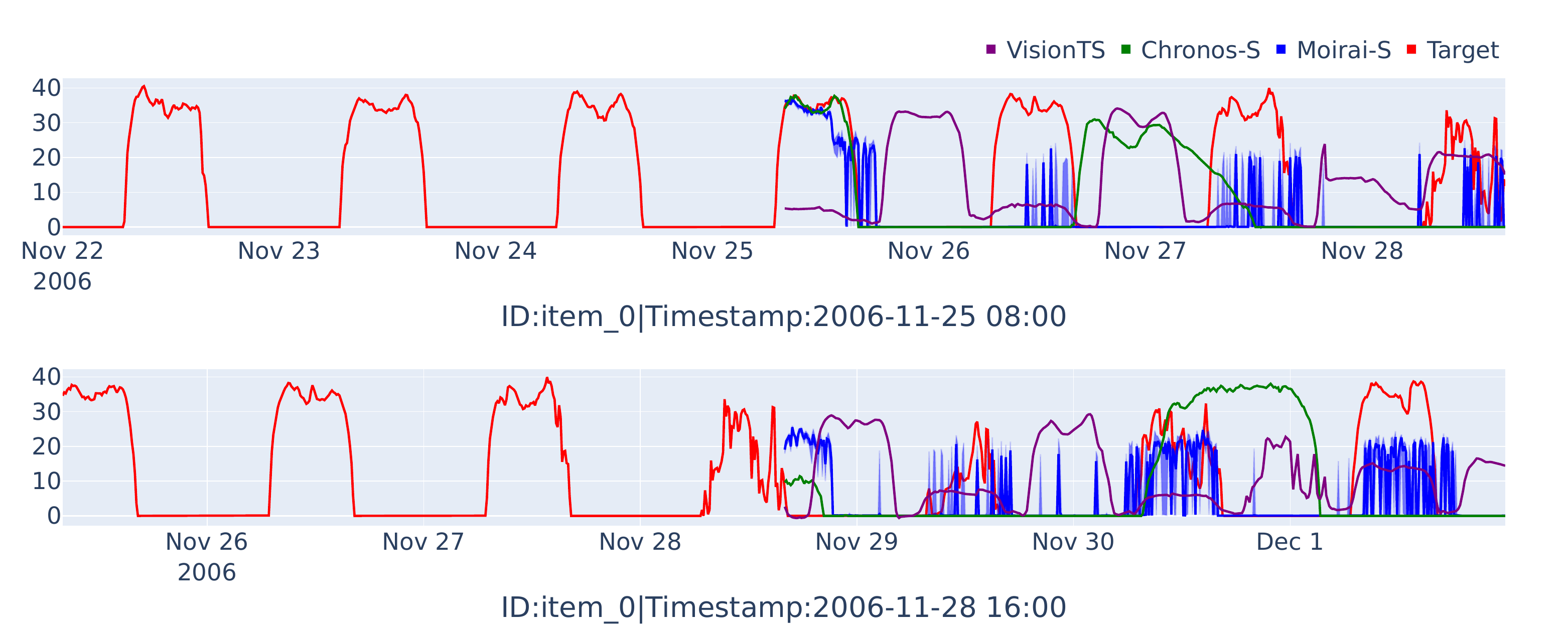}
        \caption{Foundation model forecasts sampled on \textit{Solar} ten--minutely dataset with medium prediction length.}
        \label{fig:qual4}
    \end{subfigure}
    \caption{Qualitative plots showing forecasts from various deep learning and foundation models on several time series forecasting datasets.}
    \label{fig:quals}
\end{figure}

The plots in~\Cref{fig:qual1,fig:qual2} show forecasts by deep learning models on the multivariate \textit{Bizitobs\_l2c} dataset (hourly, medium-term) and the univariate \textit{Solar} dataset (ten-minutely, medium-term). In~\Cref{fig:qual1}  the irregular patterns challenge the models, with only \patchtst{} getting close to capturing some of the regular spikes accurately. \deepar{} and \nbeats{} perform reasonably but miss key periodic spikes, while \itransformer{}, despite its multivariate capability, oversimplifies the data into a sinusoidal pattern. In~\Cref{fig:qual2}, traditional models handle seasonal data better but still tend to underpredict, with \nbeats{} producing a flat forecast in the second plot. \patchtst{} consistently outperforms others in both instances, showing robustness with both regular and irregular series, while \itransformer{} continues to underperform.

The plots in~\Cref{fig:qual3,fig:qual4} show forecasts by foundation models on two univariate datasets: \textit{M\_DENSE} (daily, long-term) and \textit{Solar} (ten-minutely, medium-term).  \Cref{fig:qual3} displays varying performance among the foundation models. \chronos{} shows a clear degradation in performance as the prediction horizon extends, struggling to maintain accuracy over time, while \visionts{} captures spikes but misaligns them. \moirai{} offers smoother, more conservative forecasts, which may result in less sensitivity to extreme events but provide more consistent alignment with the general trend. In~\Cref{fig:qual4} \visionts{} predicts seasonal peaks but with timing shifts. On the other hand, both \moirai{} and \chronos{} struggle to capture the well-spaced regularity of the data, missing key trends altogether. These poor results across all foundation models (see~\Cref{fig:qual2} vs \Cref{fig:qual4}) mirror the quantitative findings in~\Cref{sec:main_results}, \textit{i.e.} deep learning models outperform foundation models at higher frequencies. For more qualitative examples see~\Cref{app:add_quant}

\section{Conclusion}
We introduce \bench{}, a benchmark designed to evaluate time series forecasting models with diversity across four key characteristics: domain, frequency, number of variates, and prediction length. We ensure additional diversity by verifying six statistical features across temporal attributes, forecastability, and regularity. In addition to the train/test dataset, we also provide a pretraining dataset with no leakage into our evaluation set. With this, we aim to provide the necessary ground for fairly comparing different families of models, including foundation models, across a diverse benchmark. We conduct comprehensive experiments with 17 baselines encompassing statistical, deep learning, and foundation models. Leveraging our detailed taxonomy, we provide insights into each model's strengths relative to different characteristics. We also conduct a qualitative analysis highlighting failure cases in both deep learning and foundation models. \bench{} is a comprehensive benchmark with fine-grained taxonomy that we hope will accelerate the development of new foundation time series forecasting models.

\bibliography{iclr2025_conference}
\bibliographystyle{iclr2025_conference}

\appendix
\section{Experimental setup details}
\label{app:hyperparam}

\paragraph{Statistical models}
We utilize the statsforecast~\citep{garza2022statsforecast} library to implement all five statistical baselines: \naive{}, \snaive{}, \thetaf{}, \ets{} and \arima{}. Inference is performed on a CPU server equipped with 96 cores. For each dataset, a time limit of one day is set for the statistical model to complete its run. For any model that times out we halt it and replace its results with those from the \snaive{} model as a fallback. Given that some datasets in our benchmark are particularly long, we impose a maximum size constraint on each statistical baseline (set to 1000 with our time constraints), truncating the time series to this maximum size.

\paragraph{Deep learning models}

\begin{table}[!htb]
\caption{Hyperparameter search range for deep learning baselines.}
\label{tab:hparam_search}
\resizebox{\textwidth}{!}{%
\begin{tabular}{l|llllllll}
\hline
 &
  \textbf{\tide{}} &
   &
   &
   &
   &
   &
   &
   \\ \hline
\textbf{Parameters} &
  num\_layers\_encoder &
  num\_layers\_decoder &
  hidden\_dim &
  temporal\_hidden\_dim &
  decoder\_output\_dim &
  dropout\_rate &
  lr &
   \\
\textbf{Search Range} &
  {[}1,2{]} &
  {[}1,2{]} &
  {[}256,512,1024{]} &
  {[}64,128{]} &
  {[}8,16,32{]} &
  {[}0.0, 0.5{]} &
  {[}1e-5:1e-1{]} &
   \\ \hline
 &
  \textbf{\nbeats{}} &
   &
   &
   &
  \multicolumn{1}{l|}{} &
  \textbf{\patchtst{}} &
   &
   \\ \hline
\textbf{Parameters} &
  loss\_function &
  hidden\_layer\_units &
  share\_weights\_in\_stack &
  nb\_blocks\_per\_stack &
  \multicolumn{1}{l|}{lr} &
  d\_model &
  num\_encoder\_layers &
  lr \\
\textbf{Search Range} &
  {[}"mase", "mape", "smape"{]} &
  {[}256, 512, 1024, 2048{]} &
  {[}True, False{]} &
  {[}3, 4{]} &
  \multicolumn{1}{l|}{{[}1e-5:1e-1{]}} &
  {[}128, 256, 512{]} &
  {[}2, 3, 4{]} &
  {[}1e-5:1e-1{]} \\ \hline
 &
  \textbf{\itransformer{}} &
   &
  \multicolumn{1}{l|}{} &
  \textbf{\deepar{}} &
   &
  \multicolumn{1}{l|}{} &
  \textbf{\dlinear{}} &
   \\ \hline
\textbf{Parameters} &
  d\_model &
  num\_encoder\_layers &
  \multicolumn{1}{l|}{lr} &
  hidden\_size &
  num\_layers &
  \multicolumn{1}{l|}{lr} &
  lr &
   \\
\textbf{Search Range} &
  {[}128, 256, 512{]} &
  {[}2, 3, 4{]} &
  \multicolumn{1}{l|}{{[}1e-5:1e-1{]}} &
  {[}20,25,...,80{]} &
  {[}1,2,3,4{]} &
  \multicolumn{1}{l|}{{[}1e-5:1e-1{]}} &
  {[}1e-5:1e-1{]} &
   \\ \hline
 &
  \textbf{\crossformer{}} &
   &
  \multicolumn{1}{l|}{} &
  \textbf{\tft{}} &
   &
   &
   &
   \\ \hline
\textbf{Parameters} &
  d\_model &
  n\_heads &
  \multicolumn{1}{l|}{lr} &
  num\_heads &
  hidden\_dim &
  lr &
   &
   \\
\textbf{Search Range} &
  {[}64,128,256{]} &
  {[}2,4,8{]} &
  \multicolumn{1}{l|}{{[}1e-5:5e-3{]}} &
  {[}2,4,8{]} &
  {[}16,32,64{]} &
  {[}1e-5:1e-1{]} &
   &
   \\ \hline
\end{tabular}%
}
\end{table}
For all deeplearning models we either used models readily available in gluonts library~\citep{gluonts_jmlr} or we write our own wrappers. Where feasible we also add a probabilistic forecasting head to the models. Where direct probabilistic outputs are not feasible, we generate probabilistic evaluations by converting point forecasts into sample forecasts using a single sample. To identify the optimal hyperparameters, we conducted a comprehensive search across all 97 runs included in \bench{}. We employed the ray library~\citep{Moritz2017RayAD} to parallelize the search on a single GPU and used the optuna~\citep{optuna_2019} library to extend this parallelization across multiple GPU servers. We search for 15 trials for each deep learning model per each of the 97 runs. \Cref{tab:hparam_search} lists the range of parameters we search for each model. On top of the listed parameters for each model, we also search for weight decay on all runs in the range: $[1e-8:1e-2]$. For the \crossformer{} model on the long term setting of \textit{Jena Weather} dataset with both ten--minutely and hourly frequencies, we had to limit the search for \texttt{d\_model} and \texttt{n\_heads}, fixing them at 32 and 1, respectively. This adjustment was necessary because the model's attention mechanism operates across multiple variates, leading to an OOM (Out of Memory) error due to the high number of variates present in this dataset.

\paragraph{Foundation models}
For all foundation models we use their public versions available online and conduct zero-shot evaluation on our benchmark's test-split. Since Moirai~\citep{woo2024unifiedtraininguniversaltime} provides multi-patch size projections and varying context lengths. We adopt the similar approach by defining a frequency-to-patch size mapping as follows:
\begin{itemize}
    \item Yearly, Quarterly: 8
    \item Monthly: 8
    \item Weekly, Daily: 16
    \item Hourly: 32
    \item Minute-level: 32
    \item Second-level: 64
\end{itemize}
We search an optimal context length in the range $[1000, 2000, 4000, 8000]$. We used the public available \moirai{} models from the corresponding HuggingFace repos, i.e., \moiraismall{} - \url{https://huggingface.co/Salesforce/moirai-1.1-R-small}, \moiraibase{} - \url{https://huggingface.co/Salesforce/moirai-1.1-R-base}, \moirailarge{} - \url{https://huggingface.co/Salesforce/moirai-1.1-R-large}.

For \chronos{}, we mainly follow their official implementation\footnote{\url{https://github.com/amazon-science/chronos-forecasting/blob/main/scripts/evaluation/evaluate.py}} for evaluation: with the number of samples as 20. The models are loaded from the corresponding HuggingFace repos, e.g., \chronostiny{} - \url{https://huggingface.co/amazon/chronos-t5-tiny}, \chronossmall{} - \url{https://huggingface.co/amazon/chronos-t5-small}, \chronosbase - \url{https://huggingface.co/amazon/chronos-t5-base}.

For \timesfm{}, we follow their official implementation\footnote{\url{https://github.com/google-research/timesfm/blob/master/experiments/long_horizon_benchmarks/run_eval.py}} for evaluation. We set the context length for evaluation as 512 as mentioned in their paper since the maximum context length in training is 512. 
Following their default setting in their example, we keep the input patch length as 32, the output patch length as 128, the number of layers as 20, and the model dimension as 1280. 
\timesfm{} comes with only one model size, i.e., timesfm-1.0-200m, and we load the model from \url{https://huggingface.co/google/timesfm-1.0-200m}. 

For \visionts{}, we follow their official implementation\footnote{\url{https://github.com/Keytoyze/VisionTS/blob/main/eval_gluonts/run.py}} for evaluation. We set the context length as 2000, the norm constant as 0.4, the alignment constant as 0.4 according to their default settings. We use their implementation for seasonality detection to generate a candidate list and search an optimal seasonality parameter with the validation data.

\paragraph{Additional parameters and computational resources.}
All experiments are conducted on eight NVIDIA A100 GPUs. For models that has gone through training the loss function and optimizer are set following their original implementation. Additionally we set the batch size to 128 and, number of batches per epoch to 100, and finally number of epochs to 50.

\section{Details of Time Series Features}
\label{app:tsfeatures}
This section gives a detailed view of the time series features we used to analyze the test portion of our data in~\Cref{sec:train_test_data}. We use tsfeatures library~\citep{tsfeatures_nixtla} to calculate each metric. Given the scale of our dataset, we limit each time series history to the most recent 500 data points before computing the respective features. The prediction length remains faithful to the original values specified for the dataset and is not clipped.

We also acknowledge that for some overly short time series, \texttt{tsfeatures} may output \texttt{NaN} (Not a Number) values for certain features—for example, the seasonal strength of some yearly time series data. In such cases, we exclude these \texttt{NaN} values during aggregation. Below we provide specific details for each feature used:

\paragraph{Trend}

Using the STL (Seasonal and Trend decomposition using Loess) method, a time series \( x_t \) is decomposed into a trend component \( f_t \), multiple seasonal components \( s_{i,t} \) for \( i = 1, \dots, M \), and a remainder component \( e_t \):

\[
x_t = f_t + s_{1,t} + \dots + s_{M,t} + e_t,
\]

where \( M \) is the number of seasonal periods. The strength of the trend is quantified by comparing the variance of the remainder component \( e_t \) to the combined variance of the trend and remainder components. Specifically, the strength of the trend is defined as:

\[
\text{trend} = 1 - \frac{\operatorname{Var}(e_t)}{\operatorname{Var}(f_t + e_t)}.
\]

If the calculated value of \(\text{trend}\) is less than 0, it is set to 0; if it is greater than 1, it is set to 1. This measure indicates the proportion of the variability in the time series that is explained by the trend component, with values closer to 1 signifying a stronger trend.

\paragraph{Seasonal Strength}
Following the same decomposition above the strength of each seasonal component is quantified by comparing the variance of the remainder \( e_t \) to the combined variance of the seasonal component \( s_{i,t} \) and the remainder.

For each seasonal component \( s_{i,t} \), the strength of seasonality is defined as:

\[
\text{seasonal\_strength}_i = 1 - \frac{\operatorname{Var}(e_t)}{\operatorname{Var}(s_{i,t} + e_t)}.
\]

If the calculated value of \(\text{seasonal\_strength}_i\) is less than 0, it is set to 0; if it is greater than 1, it is set to 1. For non-seasonal time series, \(\text{seasonal\_strength} = 0\). This measure indicates the proportion of the variability in the time series that is explained by the \( i \)-th seasonal component, with values closer to 1 signifying stronger seasonality for that component.

\paragraph{Entropy}

Entropy is defined as the Shannon entropy of the normalized spectral density estimate \( \hat{f}(\lambda) \):

\[
\text{Entropy} = -\int_{-\pi}^{\pi} \hat{f}(\lambda) \log \hat{f}(\lambda) \, d\lambda,
\]

where \( \hat{f}(\lambda) \) is an estimate of the spectral density of the data. A lower spectral entropy indicates a higher signal-to-noise ratio, meaning the time series has more predictable patterns and is easier to forecast. Conversely, a higher spectral entropy suggests that the series is more complex and difficult to predict.

\paragraph{Hurst Exponent}

The \textit{Hurst exponent} (\textit{hurst}) is computed as \( 0.5 \) plus the maximum likelihood estimate of the fractional differencing order \( d \) by~\citet{Haslett1989}. The addition of \( 0.5 \) ensures consistency with the traditional Hurst coefficient. The values of the Hurst exponent vary between 0 and 1, with higher values indicating a smoother trend, less volatility, and less roughness.

\paragraph{Stability}
Stability measures the variability of the mean values across all tiles. It is defined as the variance of the means of the tiled windows. If the time series is divided into \( N \) tiles and \( \bar{x}_i \) represents the mean of the \( i \)-th tile, then the stability is calculated as:

\[
\text{Stability} = \operatorname{Var} \left( \bar{x}_1, \bar{x}_2, \dots, \bar{x}_N \right).
\]

A lower stability indicates that the means are consistent across tiles, suggesting a stable time series. A higher stability implies significant differences in means, indicating potential shifts or trends in the data.

\paragraph{Lumpiness}

Lumpiness assesses the variability of the variances across all tiles. It is defined as the variance of the variances of the tiled windows. Let \( s_i^2 \) denote the variance of the \( i \)-th tile. Lumpiness is then computed as:

\[
\text{Lumpiness} = \operatorname{Var} \left( s_1^2, s_2^2, \dots, s_N^2 \right).
\]

A higher lumpiness suggests that the variability within the tiles differs significantly, indicating that the time series may have periods of high and low volatility. A lower lumpiness means the variances are similar across tiles, pointing to a more homogeneous time series in terms of variability.
\section{Evaluation Metrics}
\label{app:eval_metrics}

We use two metrics to evaluate performance of forecasters: Mean Absolute Percentage Error (MAPE) for point forecasting ability and Continous Ranked Probability Score (CRPS) for probabilistic forecasting. For both metrics we use gluonts library implementation to calculate final values~\citep{alexander2020gluonts}.

\paragraph{MAPE}

MAPE is an evaluation metric used to measure the accuracy of forecasts in time series analysis. It is defined as the mean of the absolute percentage differences between the actual values \( Y_t \) and the predicted values \( \hat{Y}_t \). The formula for MAPE is:

\[
\text{MAPE} = \frac{1}{n} \sum_{t=1}^{n} \left| \frac{Y_t - \hat{Y}_t}{Y_t} \right|,
\]

where:
\begin{itemize}
    \item \( Y_t \) is the actual value at time \( t \),
    \item \( \hat{Y}_t \) is the forecasted value at time \( t \),
    \item \( n \) is the number of observations.
\end{itemize}

This metric expresses the forecast error as a percentage of the actual values, making it scale-independent and easy to interpret. However, it is sensitive to values of \( Y_t \) that are zero or close to zero, as this can lead to division by zero or inflated error percentages.

\paragraph{CRPS}

The \textit{Continuous Ranked Probability Score} (CRPS) is a metric used in probabilistic forecasting to evaluate the accuracy of predicted cumulative distribution functions (CDFs) against observed values. Given a predicted distribution with CDF \( F \) and a ground truth value \( y \), the CRPS is defined as:




\[
\text{CRPS}(F, y) = \int_{0}^{1} 2 \Lambda_{\alpha}( F^{-1}(\alpha), y ) \, d\alpha,
\]

where the quantile loss \( \Lambda_{\alpha}( q, y ) \) is defined as:

\[
\Lambda_{\alpha}( q, y ) = ( \alpha - \mathbf{1}\{ y < q \} )( y - q ).
\]

In practice, computing the CRPS integral can be computationally intensive. To address this, we approximate the CRPS using a discrete sum over a finite set of quantile levels. This approximation, often referred to as the mean weighted quantile loss~\citep{Park2021LearningQF}, is given by:

\[
\text{CRPS} \approx \frac{1}{K} \sum_{k=1}^{K} \text{wQL}[\alpha_k],
\]

where \( K \) is the number of quantile levels, and \( \{ \alpha_1, \alpha_2, \dots, \alpha_K \} \) are the selected quantile levels (e.g., \( \alpha_k = 0.1k \) for \( k = 1, 2, \dots, 9 \) when \( K = 9 \)).

The weighted quantile loss \( \text{wQL}[\alpha] \) for each quantile level \( \alpha \) is calculated as:

\[
\text{wQL}[\alpha] = 2 \frac{ \sum_{t} \Lambda_{\alpha}( \hat{q}_t(\alpha), y_t ) }{ \sum_{t} | y_t | },
\]

where:
\begin{itemize}
    \item \( \hat{q}_t(\alpha) \) is the predicted \( \alpha \)-quantile at time step \( t \),
    \item \( y_t \) is the actual observed value at time \( t \),
    \item \( \Lambda_{\alpha}( \hat{q}_t(\alpha), y_t ) \) is the quantile loss at time \( t \) for quantile level \( \alpha \).
\end{itemize}

\section{\bench{} test datasets}
\label{app:bench}
In this section we provide comprehensive list of datasets used in test portion of \bench{} along with original sources, for details regarding the pretraining portion see~\Cref{app_pretrain_data}. We utilize 10 open domain sources to curate the benchmark, here we list each one along with its properties in detail. We incorporate Jena Weather\footnote{\url{https://www.bgc-jena.mpg.de/wetter/}} dataset following \textbf{Autoformer}~\citep{Wu2021AutoformerDT}. We process BizITObs Application, Service, and L2C\footnote{\url{https://github.com/BizITObs/BizITObservabilityData/tree/main}} following the pipeline in \textbf{AutoMixer}~\citep{automixer}.These datasets consist of business and IT observability data, fusing both business KPIs and IT event channels into multivariate time series data. Within the same domain we also process Bitbrains datasets from \textbf{Grid Workloads Archive}~\citep{grid_workloads_archive}. The Restaurant data is borrowed from \textbf{Recruit Restaurant Forecasting Competition}~\citep{howard2017recruit}, The task associated with this dataset is to use reservation and visitation data to predict the total number of visitors to a restaurant for future dates. From \textbf{Informer}~\citep{zhou2021informer} we utilize ETT1 and ETT2 datasets, which denote electricity transformer temperature and serve as an indicator used in the electricity power long-term deployment. Datasets for Transport domain are extracted from \textbf{LibCity}~\citep{wang2023libcity}, which provides a collection of urban time series datasets. We utilize the solar dataset from \textbf{LSTNet}~\citep{Lai2017ModelingLA} where the task is to predict solar plant energy outputs. The second and last dataset for Sales data is by~\citet{Mancuso2020AML}. \textbf{Monash}~\citep{godahewa2021monash} is a large collection of diverse time series datasets across many domains, we choose a subset of these datasets making sure there is no leak from pretrain to test split. Finally, from \textbf{UCI ML Archive}~\citep{electricityloaddiagrams20112014_321} we use the electricity dataset which contains electricity consumption of 370 individual clients.
~\Cref{tab:dataset} lists all datasets, along with their source, frequency, prediction length and number of variates setup and presents various statistics from number of series, to series length, and also number of observations. We use last 10\% of each timeseries in the test portion of our data for testing and keep the rest for training.

\begin{table}
\caption{Individual statistics of \bench{} benchmark across all datasets.}
\label{tab:dataset}
\resizebox{\textwidth}{!}{%
\begin{tabular}{cccccccccccccccc}
\hline
\textbf{} &
  \textbf{} &
  \textbf{} &
  \textbf{} &
  \textbf{} &
  \multicolumn{3}{c}{\textbf{Series Length}} &
  \textbf{} &
  \textbf{} &
  \multicolumn{2}{c}{\textbf{Short-term}} &
  \multicolumn{2}{c}{\textbf{Med-term}} &
  \multicolumn{2}{c}{\textbf{Long-term}} \\ \hline
\textbf{Dataset} &
  \textbf{Source} &
  \textbf{Domain} &
  \textbf{Frequency} &
  \textbf{\# Series} &
  \textbf{Avg} &
  \textbf{Min} &
  \textbf{Max} &
  \textbf{\# Obs} &
  \textbf{Target Variates} &
  \textbf{Pred Length(S)} &
  \textbf{Windows} &
  \textbf{Pred Length(M)} &
  \textbf{Windows} &
  \textbf{Pred Length(L)} &
  \textbf{Windows} \\
Jena Weather &
  Autoformer~\citep{Wu2021AutoformerDT} &
  Nature &
  10T &
  1 &
  52,704 &
  52,704 &
  52,704 &
  52,704 &
  21 &
  48 &
  20 &
  480 &
  11 &
  720 &
  8 \\
Jena Weather &
  Autoformer~\citep{Wu2021AutoformerDT} &
  Nature &
  H &
  1 &
  8,784 &
  8,784 &
  8,784 &
  8,784 &
  21 &
  48 &
  19 &
  480 &
  2 &
  720 &
  2 \\
Jena Weather &
  Autoformer~\citep{Wu2021AutoformerDT} &
  Nature &
  D &
  1 &
  366 &
  366 &
  366 &
  366 &
  21 &
  30 &
  2 &
   &
   &
   &
   \\
BizITObs - Application &
  AutoMixer~\citep{automixer} &
  Web/CloudOps &
  10S &
  1 &
  8,834 &
  8,834 &
  8,834 &
  8,834 &
  2 &
  60 &
  15 &
  600 &
  2 &
  900 &
  1 \\
BizITObs - Service &
  AutoMixer~\citep{automixer} &
  Web/CloudOps &
  10S &
  21 &
  8,835 &
  8,835 &
  8,835 &
  185,535 &
  2 &
  60 &
  15 &
  600 &
  2 &
  900 &
  1 \\
BizITObs - L2C &
  AutoMixer~\citep{automixer} &
  Web/CloudOps &
  5T &
  1 &
  31,968 &
  31,968 &
  31,968 &
  31,968 &
  7 &
  48 &
  20 &
  480 &
  7 &
  720 &
  5 \\
BizITObs - L2C &
  AutoMixer~\citep{automixer} &
  Web/CloudOps &
  H &
  1 &
  2,664 &
  2,664 &
  2,664 &
  2,664 &
  7 &
  48 &
  6 &
  480 &
  1 &
  720 &
  1 \\
Bitbrains - Fast Storage &
  Grid Workloads Archive~\citep{grid_workloads_archive} &
  Web/CloudOps &
  5T &
  1,250 &
  8,640 &
  8,640 &
  8,640 &
  10,800,000 &
  2 &
  48 &
  18 &
  480 &
  2 &
  720 &
  2 \\
Bitbrains - Fast Storage &
  Grid Workloads Archive~\citep{grid_workloads_archive} &
  Web/CloudOps &
  H &
  1,250 &
  721 &
  721 &
  721 &
  901,250 &
  2 &
  48 &
  2 &
   &
   &
   &
   \\
Bitbrains - rnd &
  Grid Workloads Archive~\citep{grid_workloads_archive} &
  Web/CloudOps &
  5T &
  500 &
  8,640 &
  8,640 &
  8,640 &
  4,320,000 &
  2 &
  48 &
  18 &
  480 &
  2 &
  720 &
  2 \\
Bitbrains - rnd &
  Grid Workloads Archive~\citep{grid_workloads_archive} &
  Web/CloudOps &
  H &
  500 &
  720 &
  720 &
  720 &
  360,000 &
  2 &
  48 &
  2 &
   &
   &
   &
   \\
Restaurant &
  Recruit Rest. Comp.~\citep{howard2017recruit} &
  Sales &
  D &
  807 &
  358 &
  67 &
  478 &
  289,303 &
  1 &
  30 &
  1 &
   &
   &
   &
   \\
ETT1 &
  Informer~\citep{Zhou2020InformerBE} &
  Energy &
  15T &
  1 &
  69,680 &
  69,680 &
  69,680 &
  69,680 &
  7 &
  48 &
  20 &
  480 &
  15 &
  720 &
  10 \\
ETT1 &
  Informer~\citep{Zhou2020InformerBE} &
  Energy &
  H &
  1 &
  17,420 &
  17,420 &
  17,420 &
  17,420 &
  7 &
  48 &
  20 &
  480 &
  4 &
  720 &
  3 \\
ETT1 &
  Informer~\citep{Zhou2020InformerBE} &
  Energy &
  D &
  1 &
  725 &
  725 &
  725 &
  725 &
  7 &
  30 &
  3 &
   &
   &
   &
   \\
ETT1 &
  Informer~\citep{Zhou2020InformerBE} &
  Energy &
  W-THU &
  1 &
  103 &
  103 &
  103 &
  103 &
  7 &
  8 &
  2 &
   &
   &
   &
   \\
ETT2 &
  Informer~\citep{Zhou2020InformerBE} &
  Energy &
  15T &
  1 &
  69,680 &
  69,680 &
  69,680 &
  69,680 &
  7 &
  48 &
  20 &
  480 &
  15 &
  720 &
  10 \\
ETT2 &
  Informer~\citep{Zhou2020InformerBE} &
  Energy &
  H &
  1 &
  17,420 &
  17,420 &
  17,420 &
  17,420 &
  7 &
  48 &
  20 &
  480 &
  4 &
  720 &
  3 \\
ETT2 &
  Informer~\citep{Zhou2020InformerBE} &
  Energy &
  D &
  1 &
  725 &
  725 &
  725 &
  725 &
  7 &
  30 &
  3 &
   &
   &
   &
   \\
ETT2 &
  Informer~\citep{Zhou2020InformerBE} &
  Energy &
  W-THU &
  1 &
  103 &
  103 &
  103 &
  103 &
  7 &
  8 &
  2 &
   &
   &
   &
   \\
Loop Seattle &
  LibCity~\citep{wang2023libcity} &
  Transport &
  5T &
  323 &
  105,120 &
  105,120 &
  105,120 &
  33,953,760 &
  1 &
  48 &
  20 &
  480 &
  20 &
  720 &
  15 \\
Loop Seattle &
  LibCity~\citep{wang2023libcity} &
  Transport &
  H &
  323 &
  8,760 &
  8,760 &
  8,760 &
  2,829,480 &
  1 &
  48 &
  19 &
  480 &
  2 &
  720 &
  2 \\
Loop Seattle &
  LibCity~\citep{wang2023libcity} &
  Transport &
  D &
  323 &
  365 &
  365 &
  365 &
  117,895 &
  1 &
  30 &
  2 &
   &
   &
   &
   \\
SZ-Taxi &
  LibCity~\citep{wang2023libcity} &
  Transport &
  15T &
  156 &
  2,976 &
  2,976 &
  2,976 &
  464,256 &
  1 &
  48 &
  7 &
  480 &
  1 &
  720 &
  1 \\
SZ-Taxi &
  LibCity~\citep{wang2023libcity} &
  Transport &
  H &
  156 &
  744 &
  744 &
  744 &
  116,064 &
  1 &
  48 &
  2 &
   &
   &
   &
   \\
M\_DENSE &
  LibCity~\citep{wang2023libcity} &
  Transport &
  H &
  30 &
  17,520 &
  17,520 &
  17,520 &
  525,600 &
  1 &
  48 &
  20 &
  480 &
  4 &
  720 &
  3 \\
M\_DENSE &
  LibCity~\citep{wang2023libcity} &
  Transport &
  D &
  30 &
  730 &
  730 &
  730 &
  21,900 &
  1 &
  30 &
  3 &
   &
   &
   &
   \\
Solar &
  LSTNet~\citep{Lai2017ModelingLA} &
  Energy &
  10T &
  137 &
  52,560 &
  52,560 &
  52,560 &
  7,200,720 &
  1 &
  48 &
  20 &
  480 &
  11 &
  720 &
  8 \\
Solar &
  LSTNet~\citep{Lai2017ModelingLA} &
  Energy &
  H &
  137 &
  8,760 &
  8,760 &
  8,760 &
  1,200,120 &
  1 &
  48 &
  19 &
  480 &
  2 &
  720 &
  2 \\
Solar &
  LSTNet~\citep{Lai2017ModelingLA} &
  Energy &
  D &
  137 &
  365 &
  365 &
  365 &
  50,005 &
  1 &
  30 &
  2 &
   &
   &
   &
   \\
Solar &
  LSTNet~\citep{Lai2017ModelingLA} &
  Energy &
  W-FRI &
  137 &
  52 &
  52 &
  52 &
  7,124 &
  1 &
  8 &
  1 &
   &
   &
   &
   \\
Hierarchical Sales &
  \citet{Mancuso2020AML} &
  Sales &
  D &
  118 &
  1,825 &
  1,825 &
  1,825 &
  215,350 &
  1 &
  30 &
  7 &
   &
   &
   &
   \\
Hierarchical Sales &
  \citet{Mancuso2020AML} &
  Sales &
  W-WED &
  118 &
  260 &
  260 &
  260 &
  30,680 &
  1 &
  8 &
  4 &
   &
   &
   &
   \\
M4 Yearly &
  Monash~\citep{godahewa2021monash} &
  Econ/Fin &
  A-DEC &
  22,974 &
  37 &
  19 &
  284 &
  845,109 &
  1 &
  6 &
  1 &
   &
   &
   &
   \\
M4 Quarterly &
  Monash~\citep{godahewa2021monash} &
  Econ/Fin &
  Q-DEC &
  24,000 &
  100 &
  24 &
  874 &
  2,406,108 &
  1 &
  8 &
  1 &
   &
   &
   &
   \\
M4 Monthly &
  Monash~\citep{godahewa2021monash} &
  Econ/Fin &
  M &
  48,000 &
  234 &
  60 &
  2,812 &
  11,246,411 &
  1 &
  18 &
  1 &
   &
   &
   &
   \\
M4 Weekly &
  Monash~\citep{godahewa2021monash} &
  Econ/Fin &
  W-SUN &
  359 &
  1,035 &
  93 &
  2,610 &
  371,579 &
  1 &
  13 &
  1 &
   &
   &
   &
   \\
M4 Daily &
  Monash~\citep{godahewa2021monash} &
  Econ/Fin &
  D &
  4,227 &
  2,371 &
  107 &
  9,933 &
  10,023,836 &
  1 &
  14 &
  1 &
   &
   &
   &
   \\
M4 Hourly &
  Monash~\citep{godahewa2021monash} &
  Econ/Fin &
  H &
  414 &
  902 &
  748 &
  1,008 &
  373,372 &
  1 &
  48 &
  2 &
   &
   &
   &
   \\
Hospital &
  Monash~\citep{godahewa2021monash} &
  Healthcare &
  M &
  767 &
  84 &
  84 &
  84 &
  64,428 &
  1 &
  12 &
  1 &
   &
   &
   &
   \\
COVID Deaths &
  Monash~\citep{godahewa2021monash} &
  Healthcare &
  D &
  266 &
  212 &
  212 &
  212 &
  56,392 &
  1 &
  30 &
  1 &
   &
   &
   &
   \\
US Births &
  Monash~\citep{godahewa2021monash} &
  Healthcare &
  D &
  1 &
  7,305 &
  7,305 &
  7,305 &
  7,305 &
  1 &
  30 &
  20 &
   &
   &
   &
   \\
US Births &
  Monash~\citep{godahewa2021monash} &
  Healthcare &
  W-TUE &
  1 &
  1,043 &
  1,043 &
  1,043 &
  1,043 &
  1 &
  8 &
  14 &
   &
   &
   &
   \\
US Births &
  Monash~\citep{godahewa2021monash} &
  Healthcare &
  M &
  1 &
  240 &
  240 &
  240 &
  240 &
  1 &
  12 &
  2 &
   &
   &
   &
   \\
Saugeen &
  Monash~\citep{godahewa2021monash} &
  Nature &
  D &
  1 &
  23,741 &
  23,741 &
  23,741 &
  23,741 &
  1 &
  30 &
  20 &
   &
   &
   &
   \\
Saugeen &
  Monash~\citep{godahewa2021monash} &
  Nature &
  W-THU &
  1 &
  3,391 &
  3,391 &
  3,391 &
  3,391 &
  1 &
  8 &
  20 &
   &
   &
   &
   \\
Saugeen &
  Monash~\citep{godahewa2021monash} &
  Nature &
  M &
  1 &
  780 &
  780 &
  780 &
   &
  1 &
  12 &
  7 &
   &
   &
   &
   \\
Temperature Rain &
  Monash~\citep{godahewa2021monash} &
  Nature &
  D &
  32,072 &
  725 &
  725 &
  725 &
  780 &
  1 &
  30 &
  3 &
   &
   &
   &
   \\
KDD Cup 2018 &
  Monash~\citep{godahewa2021monash} &
  Nature &
  H &
  270 &
  10,898 &
  9,504 &
  10,920 &
  2,942,364 &
  1 &
  48 &
  20 &
  480 &
  2 &
  720 &
  2 \\
KDD Cup 2018 &
  Monash~\citep{godahewa2021monash} &
  Nature &
  D &
  270 &
  455 &
  396 &
  455 &
  122,791 &
  1 &
  30 &
  2 &
   &
   &
   &
   \\
Car Parts &
  Monash~\citep{godahewa2021monash} &
  Sales &
  M &
  2,674 &
  51 &
  51 &
  51 &
  136,374 &
  1 &
  12 &
  1 &
   &
   &
   &
   \\
Electricity &
  UCI ML Archive~\citep{electricityloaddiagrams20112014_321} &
  Energy &
  15T &
  370 &
  140,256 &
  140,256 &
  140,256 &
  51,894,720 &
  1 &
  48 &
  20 &
  480 &
  20 &
  720 &
  20 \\
Electricity &
  UCI ML Archive~\citep{electricityloaddiagrams20112014_321} &
  Energy &
  H &
  370 &
  35,064 &
  35,064 &
  35,064 &
  12,973,680 &
  1 &
  48 &
  20 &
  480 &
  8 &
  720 &
  5 \\
Electricity &
  UCI ML Archive~\citep{electricityloaddiagrams20112014_321} &
  Energy &
  D &
  370 &
  1,461 &
  1,461 &
  1,461 &
  540,570 &
  1 &
  30 &
  5 &
   &
   &
   &
   \\
Electricity &
  UCI ML Archive~\citep{electricityloaddiagrams20112014_321} &
  Energy &
  W-FRI &
  370 &
  208 &
  208 &
  208 &
  76,960 &
  1 &
  8 &
  3 &
   &
   &
   &
   \\ \hline
\end{tabular}%
}
\end{table}

\section{\bench{} pre-training datasets}
\label{app_pretrain_data}
The pre-training split of \bench{} is constructed based on LOTSA \citep{woo2024unifiedtraininguniversaltime}, and we excluded certain datasets from it to form part of the evaluation set, making it more diverse and balanced.

Here is a brief discussion on each of the used sources:~\textbf{BuildingsBench} \citep{emami2023buildingsbench} compiled datasets on residential and commercial building energy consumption. \textbf{ClimateLearn} \citep{nguyen2023climatelearn} offered time series of various climate-related variables, including temperature, humidity, and multiple pressure levels. \textbf{CloudOps TSF} \citep{woo2023pushing} introduced large-scale CloudOps time series datasets that capture key variables such as CPU and memory utilization. \textbf{GluonTS} \citep{alexander2020gluonts} provided a variety of datasets commonly used in time series forecasting. \textbf{LargeST} \citep{liu2023largest} sourced from the California Department of Transportation Performance Measurement System (PeMS) to date, which is widely used for traffic forecasting. \textbf{LibCity} \citep{wang2023libcity} provided a collection urban spatio-temporal datasets. \textbf{SubseasonalClimateUSA} \citep{mouatadid2023subseasonalclimateusa} provided climate time series data at daily level. \textbf{ProEnFo} \citep{wang2023proenfo} introduced a range of datasets for load forecasting which include various covariates such as temperature, humidity, and wind speed. \textbf{Monash} \citep{godahewa2021monash} is a large collection of diverse time series datasets, the most popular source for building time series foundation models. \textbf{LOTSA$\_$Others} \citep{woo2024unifiedtraininguniversaltime} are complementary datasets collected by LOTSA to enhance the diversity.

The complete list of pre-training datasets and their respective sources, key properties are provided in ~\Cref{tab:pretraining}.


\begin{table}[ht]
  \centering
  \caption{Pretraining datasets and their key properties.}
  \resizebox{\textwidth}{!}{
    \begin{tabular}{lccccccc}
    \toprule
    \textbf{Dataset} & \textbf{Source} & \textbf{Domain} &\textbf{Frequency} & 
    \textbf{\# Time Series} & \textbf{\# Targets} & \textbf{\# Covariates} & 
    \textbf{\# Obs.} \\
    \midrule
    BDG-2 Panther & BuildingsBench~\citep{emami2023buildingsbench} & Energy   & H     & 105   & 1     & 0     & 919,800 \\
    BDG-2 Fox & BuildingsBench~\citep{emami2023buildingsbench} & Energy  & H     & 135   & 1     & 0     & 2,324,568 \\
    BDG-2 Rat & BuildingsBench~\citep{emami2023buildingsbench} & Energy  & H     & 280   & 1     & 0     & 4,728,288 \\
    BDG-2 Bear & BuildingsBench~\citep{emami2023buildingsbench} & Energy  & H     & 91    & 1     & 0     & 1,482,312 \\
    Low Carbon London & BuildingsBench~\citep{emami2023buildingsbench} & Energy  & H     & 713   & 1     & 0     & 9,543,348 \\
    SMART & BuildingsBench~\citep{emami2023buildingsbench} & Energy  & H     & 5     & 1     & 0     & 95,709 \\
    IDEAL & BuildingsBench~\citep{emami2023buildingsbench} & Energy  & H     & 219   & 1     & 0     & 1,265,672 \\
    Sceaux & BuildingsBench~\citep{emami2023buildingsbench} & Energy  & H     & 1     & 1     & 0     & 34,223 \\
    Borealis & BuildingsBench~\citep{emami2023buildingsbench} & Energy  & H     & 15    & 1     & 0     & 83,269 \\
    Buildings900K & BuildingsBench~\citep{emami2023buildingsbench} & Energy  & H     & 1,792,328 & 1     & 0     & 15,702,590,000 \\ 
    CMIP6 & ClimateLearn~\citep{nguyen2023climatelearn} & Climate  & 6H    & 1,351,680 & 53    & 0     & 1,973,453,000 \\
    ERA5  & ClimateLearn~\citep{nguyen2023climatelearn} & Climate  & H     & 245,760 & 45    & 0     & 2,146,959,000 \\
    Azure VM Traces 2017 & CloudOpsTSF~\citep{woo2023pushing} & CloudOps  &5T    & 159,472 & 1     & 2     & 885,522,908 \\
    Borg Cluster Data 2011 & CloudOpsTSF~\citep{woo2023pushing} & CloudOps  & 5T    & 143,386 & 2     & 5     & 537,552,854 \\
    Alibaba Cluster Trace 2018 & CloudOpsTSF~\citep{woo2023pushing} & CloudOps  & 5T    & 58,409 & 2     & 6     & 95,192,530 \\
    Taxi  & GluonTS~\citep{alexander2020gluonts} & Transport  & 30T   & 67,984 & 1     & 0     & 54,999,060 \\
    Uber TLC Daily & GluonTS~\citep{alexander2020gluonts} & Transport  & D     & 262   & 1     & 0     & 47,087 \\
    Uber TLC Hourly & GluonTS~\citep{alexander2020gluonts} & Transport  & H     & 262   & 1     & 0     & 1,129,444 \\
    Wiki-Rolling & GluonTS~\citep{alexander2020gluonts} & Web    & D     & 47,675 & 1     & 0     & 40,619,100 \\
    M5    & GluonTS~\citep{alexander2020gluonts} & Sales  & D     & 30,490 & 1     & 0     & 58,327,370 \\
    LargeST & LargeST~\citep{liu2023largest} & Transport  & 5T    & 42,333 & 1     & 0     & 4,452,510,528 \\
    PEMS03 &LibCity~\citep{wang2023libcity} & Transport & 5T    & 358   & 1     & 0     & 9,382,464 \\
    PEMS04 &LibCity~\citep{wang2023libcity} & Transport & 5T    & 307   & 3     & 0     & 5,216,544 \\
    PEMS07 &LibCity~\citep{wang2023libcity} & Transport & 5T    & 883   & 1     & 0     & 24,921,792 \\
    PEMS08 &LibCity~\citep{wang2023libcity} & Transport & 5T    & 170   & 3     & 0     & 3,035,520 \\
    PEMS Bay &LibCity~\citep{wang2023libcity} & Transport & 5T    & 325   & 1     & 0     & 16,937,700 \\
    Los-Loop &LibCity~\citep{wang2023libcity} & Transport & 5T    & 207   & 1     & 0     & 7,094,304 \\
    Beijing Subway&LibCity~\citep{wang2023libcity}  & Transport & 30T   & 276   & 2     & 11    & 248,400 \\
    SHMetro &LibCity~\citep{wang2023libcity}  & Transport & 15T   & 288   & 2     & 0     & 1,934,208 \\
    HZMetro &LibCity~\citep{wang2023libcity} & Transport & 15T   & 80    & 2     & 0     & 146,000 \\
    Q-Traffic &LibCity~\citep{wang2023libcity}  & Transport & 15T   & 45,148 & 1     & 0     & 264,386,688 \\
    Subseasonal & SubseasonalClimateUSA~\citep{mouatadid2023subseasonalclimateusa} & Climate & D     & 862   & 4     & 0     & 14,097,148 \\
    Subseasonal Precipitation& SubseasonalClimateUSA ~\citep{mouatadid2023subseasonalclimateusa} & Climate & D     & 862   & 1     & 0     & 9,760,426 \\
    Covid19 Energy & ProEnFo~\citep{wang2023proenfo} & Energy & H     & 1     & 1     & 6     & 31,912 \\
    GEF12 & ProEnFo ~\citep{wang2023proenfo} & Energy & H     & 20    & 1     & 1     & 788,280 \\
    GEF14 & ProEnFo~\citep{wang2023proenfo} & Energy & H     & 1     & 1     & 1     & 17,520 \\
    GEF17 & ProEnFo~\citep{wang2023proenfo} & Energy & H     & 8     & 1     & 1     & 140,352 \\
    PDB   & ProEnFo~\citep{wang2023proenfo} & Energy & H     & 1     & 1     & 1     & 17,520 \\
    Spanish & ProEnFo~\citep{wang2023proenfo} & Energy & H     & 1     & 1     & 1     & 35,064 \\
    BDG-2 Hog & ProEnFo~\citep{wang2023proenfo} & Energy & H     & 24    & 1     & 5     & 421,056 \\
    BDG-2 Bull & ProEnFo~\citep{wang2023proenfo} & Energy & H     & 41    & 1     & 3     & 719,304 \\
    BDG-2 Cockatoo & ProEnFo~\citep{wang2023proenfo} & Energy & H     & 1     & 1     & 5     & 17,544 \\
    ELF  & ProEnFo~\citep{wang2023proenfo}  & Energy & H     & 1     & 1     & 0     & 21,792 \\
    London Smart Meters & Monash~\citep{godahewa2021monash} & Energy & 30T   & 5,520 & 1     & 0     & 166,238,880 \\
    Wind Farms & Monash~\citep{godahewa2021monash} & Energy   & T     & 337   & 1     & 0     & 172,165,370 \\
    Wind Power  & Monash~\citep{godahewa2021monash} & Energy & 4S    & 1     & 1     & 0     & 7,397,147 \\
    Solar Power  & Monash~\citep{godahewa2021monash} & Energy & 4S    & 1     & 1     & 0     & 7,397,222 \\
    Oikolab Weather  & Monash~\citep{godahewa2021monash} & Climate & H     & 8     & 1     & 0     & 800,456 \\
    Elecdemand & Monash~\citep{godahewa2021monash} & Energy & 30T   & 1     & 1     & 0     & 17,520 \\
    Covid Mobility  & Monash~\citep{godahewa2021monash} & Transport & D     & 362   & 1     & 0     & 148,602 \\
    Kaggle Web Traffic Weekly & Monash~\citep{godahewa2021monash} & Web   & W     & 145,063 & 1     & 0     & 16,537,182 \\
    Extended Web Traffic & Monash~\citep{godahewa2021monash} & Web   & D     & 145,063 & 1     & 0     & 370,926,091 \\
    M1 Yearly & Monash~\citep{godahewa2021monash} & Econ/Fin & Y     & 106   & 1     & 0     & 3,136 \\
    M1 Quarterly & Monash~\citep{godahewa2021monash}& Econ/Fin & Q     & 198   & 1     & 0     & 9,854 \\
    M1 Monthly & Monash~\citep{godahewa2021monash} & Econ/Fin & M     & 617   & 1     & 0     & 44,892 \\
    M3 Yearly & Monash~\citep{godahewa2021monash} & Econ/Fin & Y     & 645   & 1     & 0     & 18,319 \\
    M3 Quarterly & Monash~\citep{godahewa2021monash}  & Econ/Fin & Q     & 756   & 1     & 0     & 37,004 \\
    M3 Monthly & Monash~\citep{godahewa2021monash} & Econ/Fin & M     & 1,428 & 1     & 0     & 141,858 \\
    M3 Other & Monash~\citep{godahewa2021monash} & Econ/Fin & Q     & 174   & 1     & 0     & 11,933 \\
    NN5 Daily & Monash~\citep{godahewa2021monash} & Econ/Fin & D     & 111   & 1     & 0     & 81,585 \\
    NN5 Weekly & Monash~\citep{godahewa2021monash} & Econ/Fin & W     & 111   & 1     & 0     & 11,655 \\
    Tourism Yearly & Monash~\citep{godahewa2021monash} & Econ/Fin & Y     & 419   & 1     & 0     & 11,198 \\
    Tourism Quarterly & Monash~\citep{godahewa2021monash} & Econ/Fin & Q     & 427   & 1     & 0     & 39,128 \\
    Tourism Monthly & Monash~\citep{godahewa2021monash}& Econ/Fin & M     & 366   & 1     & 0     & 100,496 \\
    CIF 2016 & Monash~\citep{godahewa2021monash} & Econ/Fin  & M     & 72   & 1     & 0     & 6,334 \\
    Traffic Weekly & Monash~\citep{godahewa2021monash} & Transport & W     & 862   & 1     & 0     & 82,752 \\
    Traffic Hourly & Monash~\citep{godahewa2021monash} & Transport & H     & 862   & 1     & 0     & 14,978,112 \\
    Australian Electricity Demand & Monash~\citep{godahewa2021monash} & Energy & 30T   & 5     & 1     & 0     & 1,153,584 \\
    Rideshare & Monash~\citep{godahewa2021monash}  & Transport & H     & 2,304 & 1     & 0     & 859,392 \\
    Sunspot & Monash~\citep{godahewa2021monash}  & Nature & D     & 1     & 1     & 0     & 73,894 \\
    Vehicle Trips & Monash~\citep{godahewa2021monash} & Transport & D     & 329   & 1     & 0     & 32,512 \\
    Weather & Monash~\citep{godahewa2021monash} & Climate & D     & 3,010 & 1     & 0     & 42,941,700 \\
    FRED MD & Monash~\citep{godahewa2021monash} & Econ/Fin & M     & 107   & 1     & 0     & 76,612 \\
    Pedestrian Counts & Monash~\citep{godahewa2021monash} & Transport & H     & 66    & 1     & 0     & 3,130,762 \\
    Bitcoin & Monash~\citep{godahewa2021monash} & Econ/Fin & D     & 18    & 1     & 0     & 74,824 \\
     KDD Cup 2022 &    LOTSA$\_$Others~\citep{woo2024unifiedtraininguniversaltime}   & Energy & 10T   & 134   & 1     & 9     & 4,727,519 \\
    GoDaddy & LOTSA$\_$Others~\citep{woo2024unifiedtraininguniversaltime} & Econ/Fin & M     & 3,135 & 2     & 0     & 128,535 \\
    Favorita Sales & LOTSA$\_$Others~\citep{woo2024unifiedtraininguniversaltime} & Sales & D     & 111,840 & 1     & 0     & 139,179,538 \\
    Favorita Transactions & LOTSA$\_$Others~\citep{woo2024unifiedtraininguniversaltime} & Sales & D     & 54    & 1     & 0     & 84,408 \\
    China Air Quality &   LOTSA$\_$Others~\citep{woo2024unifiedtraininguniversaltime}   & Nature & H     & 437   & 6     & 0     & 5,739,234 \\
    Beijing Air Quality & LOTSA$\_$Others~\citep{woo2024unifiedtraininguniversaltime}   & Nature & H     & 12    & 11    & 0     & 420,768 \\
    Residential Load Power &   LOTSA$\_$Others~\citep{woo2024unifiedtraininguniversaltime}    & Energy & T     & 271   & 3     & 0     & 145,994,559 \\
    Residential PV Power &   LOTSA$\_$Others~\citep{woo2024unifiedtraininguniversaltime}    & Energy & T     & 233   & 3     & 0     & 125,338,950 \\
    CDC Fluview ILINet & LOTSA$\_$Others~\citep{woo2024unifiedtraininguniversaltime}   & Healthcare & W     & 75    & 5     & 0     & 63,903 \\
    CDC Fluview WHO NREVSS & LOTSA$\_$Others~\citep{woo2024unifiedtraininguniversaltime}   & Healthcare & W     & 74    & 4     & 0     & 41,760 \\
    Project Tycho & LOTSA$\_$Others~\citep{woo2024unifiedtraininguniversaltime} & Healthcare & W     & 1,258 & 1     & 0     & 1,377,707 \\
    \bottomrule
    \end{tabular}}%
  \label{tab:pretraining}%
\end{table}%

\section{Finegrained Results}
\label{app:finegrained_results}

\subsection{Results with all models}
In this section, we present results for all models, including those omitted from the main paper in section~\Cref{sec:main_results} due to space constraints. The results are displayed in the same aggregated form through~\Cref{tab:full_domain_results,tab:full_term_length_results,tab:full_frequency_results,tab:full_variate_results,tab:full_results}. Furthermore, we provide non-aggregated results across all dataset, term and frequency combinations in~\Cref{tab:all_results,tab:all_results2,tab:all_results3}.

\subsection{Data Leakage Effect in Foundation Models}
\label{app:leakage}
Due to the use of different pre-training datasets by various foundation models, there is a partial data leakage issue when these models are evaluated against our \bench{}. To ensure a fair comparison, we pre-trained a new series of \moirai{} models on our \bench{} pre-training data and report the results from these models in the main paper.  To further examine the impact of data leakage on foundation models, we also include the performance of the original \moirai{} models on our benchmark, both to facilitate replicability and to demonstrate the misleading effects that leakage can introduce.  We call the original model \omoirai{} and use the abbreviation \tomoirai{}. The results for all datasets affected by leakage are provided in ~\Cref{tab:leak_results}. In most cases, data leakage from training sets can  boost performance on the corresponding test sets, with this effect becoming more pronounced as prediction length increases. This finding highlights the capacity of foundation models to memorize training data and underscores the critical importance of preventing data leakage when comparing time series foundation models on public benchmarks.

\subsection{Additional Qualitative Examples}
\label{app:add_quant}
\begin{figure}[htb!]
    \centering
    \begin{subfigure}{\textwidth}
    \includegraphics[width=\textwidth]{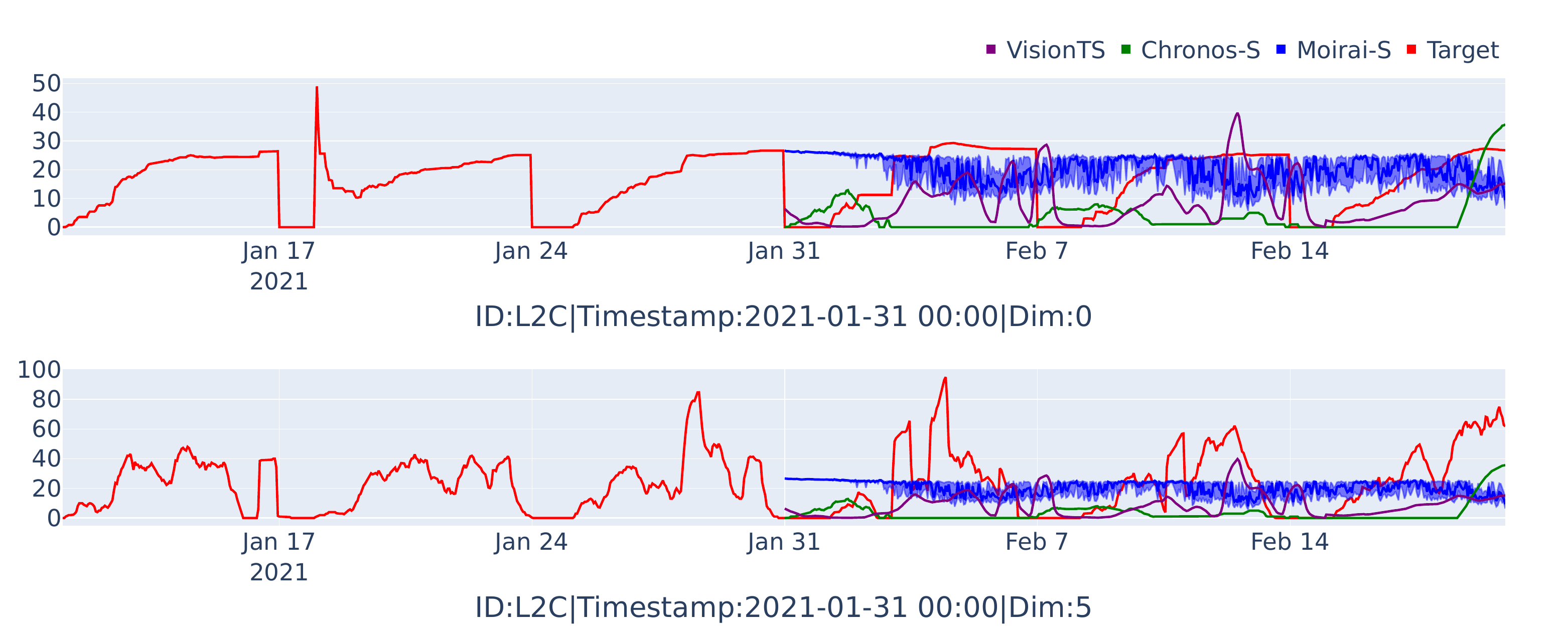}
        \caption{Foundation model forecasts sampled on \textit{Bizitobs\_l2c} hourly dataset with medium prediction length.}
        \label{fig:app_qual1}
    \end{subfigure}%
    \hfill
    \begin{subfigure}{\textwidth}
        \includegraphics[width=\textwidth]{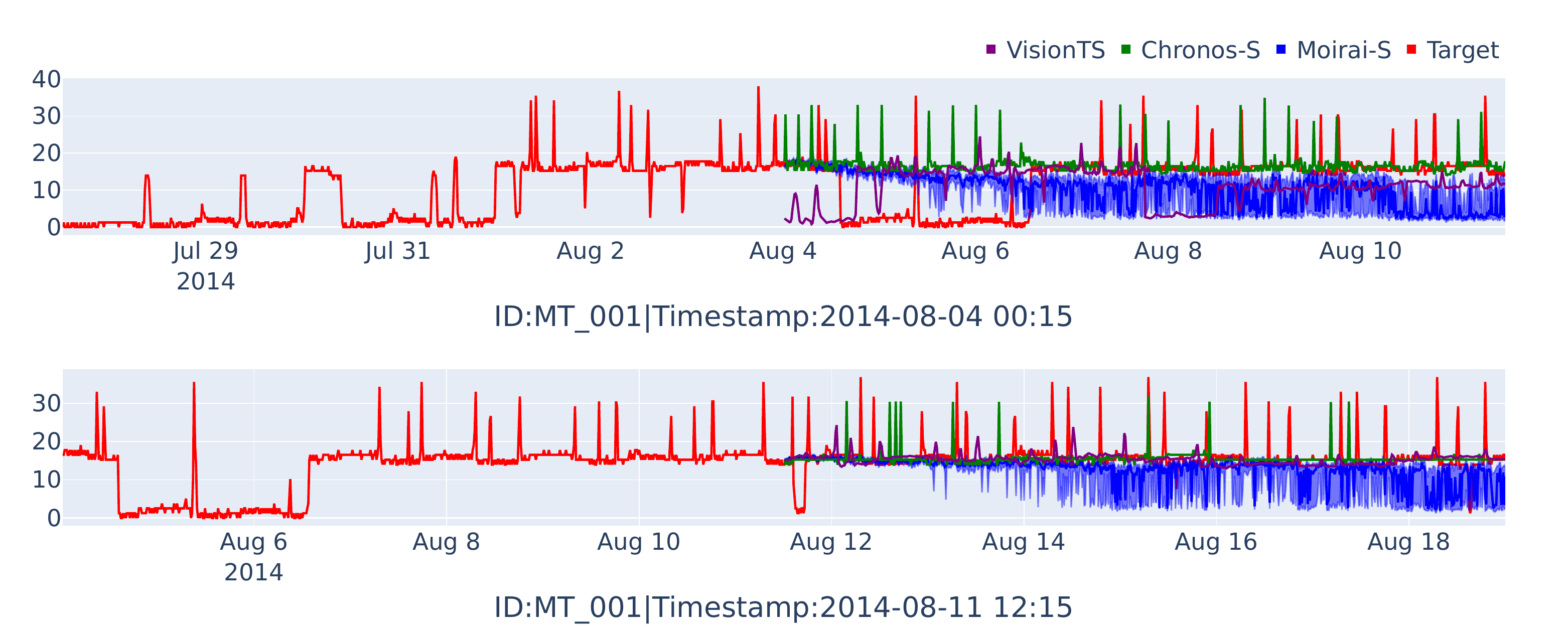}
        \caption{Foundation model forecasts sampled on \textit{Electricity} 15--minutely dataset with long prediction length.}
        \label{fig:app_qual2}
    \end{subfigure}%
    \hfill
    \begin{subfigure}{\textwidth}
        \includegraphics[width=\textwidth]{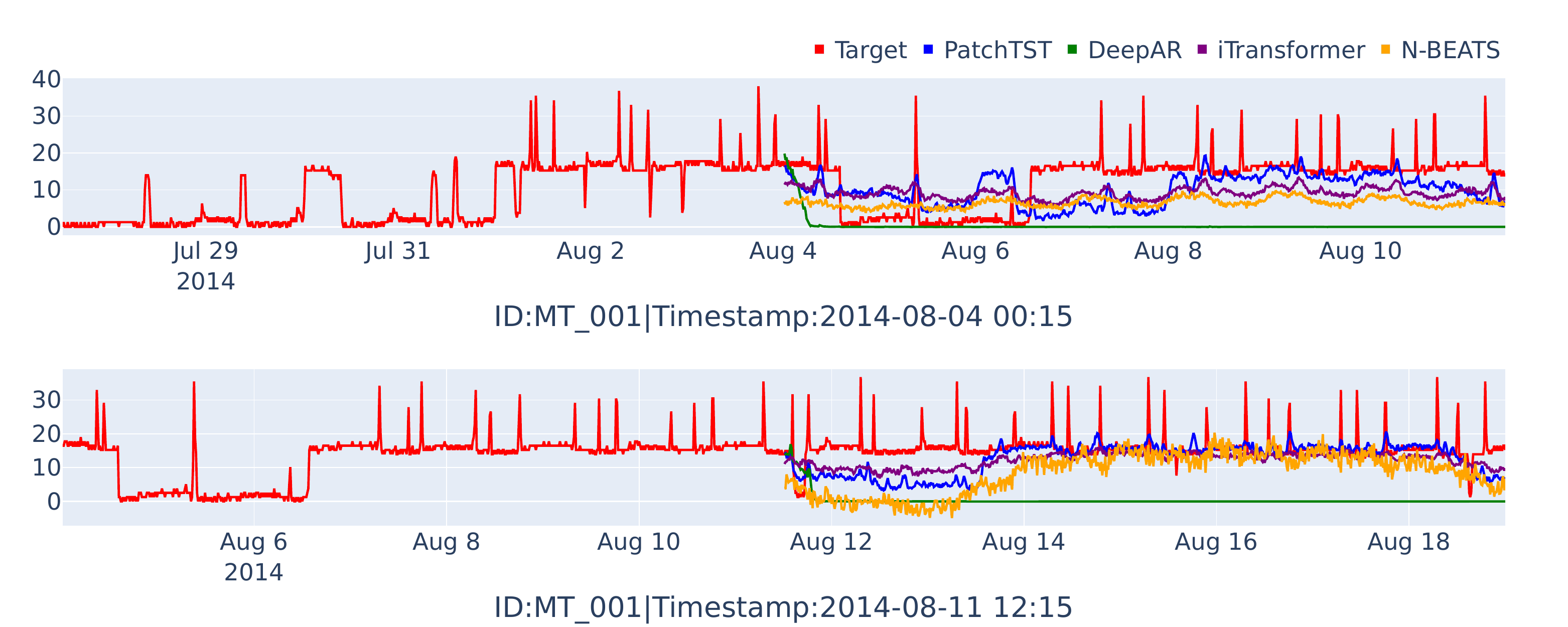}
        \caption{Deep learning model forecasts sampled on \textit{Electricity} 15--minutely dataset with long prediction length.}
        \label{fig:app_qual3}
    \end{subfigure}
    \caption{Qualitative plots showing forecasts from various deep learning and foundation models on several time series forecasting datasets.}
    \label{fig:app_quals}
\end{figure}

In addition to the four examples shared in the main paper, we present three additional qualitative examples in~\Cref{fig:app_quals}. ~\Cref{fig:app_qual1} illustrates the forecasts of foundation models on the \textit{Bizitobs\_l2c} dataset (hourly, medium-term). Similar to previous observations, \chronos{} forecasts tend to degrade over longer time horizons. Unlike in earlier scenarios, \moirai{} also shows poor performance on this dataset, missing all the regular peaks and troughs. In contrast, the \visionts{} model provides the forecast closest to the ground truth. ~\Cref{fig:app_qual2} presents forecasts for a higher-frequency dataset: \textit{Electricity} (15-minute intervals, long-term). Notably, \chronos{} excels on this dataset, showing better consistency than both \moirai{} and \visionts{}. While \moirai{} performs reasonably well, it tends to predict some stationary changes (see the rightmost side of the upper plot) that are not aligned with the ground truth data. In contrast, \visionts{} repeats a mistake observed in earlier datasets by predicting shifted peaks. The final plots in~\Cref{fig:app_qual3} display forecasts from deep learning models on the same \textit{Electricity} dataset (15-minute intervals, long-term). Compared to the foundation models, these deep learning models demonstrate poorer performance. Notably, the model that differs most by its forecast is \deepar{}, which quickly flattens at the beginning of the prediction—a phenomenon also observed with another deep learning model, \nbeats{}, in the \textit{Solar} dataset example in~\Cref{fig:qual2}.
\begin{table}
\Large

\caption{Results on \bench{} with all models aggregated by domain. The best results across each row are \textbf{bolded}, while second best results are \underline{underlined}.}
\label{tab:full_domain_results}
\resizebox{\textwidth}{!}{%
\begin{tabular}{ll|lllll|llllllll|llllllll|l}
\toprule
Domain & Metric & \tnaive{} & \tsnaive{} & \tarima{} & \tthetaf{} & \tets{} & \tdeepar{} & \ttft{} & \ttide{} & \tnbeats{} & \tpatchtst{} & \titransformer{} & \tdlinear{} & \tcrossformer{} & \ttimesfm{} & \tvisionts{} & \tchronossmall{} & \tchronosbase{} & \tchronoslarge{} & \tmoiraismall{} & \tmoiraibase{} & \tmoirailarge{} & Best \\
\hline
\multirow{1}{*}{Econ/Fin}& MAPE & 1.22 & 1.00 & $9.37e^{-1}$ & 1.01 & 1.07 & 1.23 & 1.14 & 1.15 & $8.63e^{-1}$ & $9.46e^{-1}$ & 1.02 & 1.11 & $3.43e^{1}$ & $8.41e^{-1}$ & $9.93e^{-1}$ & \underline{$8.02e^{-1}$} & $8.02e^{-1}$ & \boldmath{$7.99e^{-1}$} & 1.04 & 1.27 & $8.80e^{-1}$ & \tchronoslarge{} \\
 & CRPS & 1.36 & 1.00 & $8.25e^{-1}$ & $8.46e^{-1}$ & $9.91e^{-1}$ & 1.41 & $8.48e^{-1}$ & 1.09 & $9.86e^{-1}$ & $8.15e^{-1}$ & $8.55e^{-1}$ & 1.13 & $1.27e^{2}$ & \boldmath{$7.29e^{-1}$} & 1.06 & $7.78e^{-1}$ & $7.68e^{-1}$ & $7.74e^{-1}$ & $8.53e^{-1}$ & 1.04 & \underline{$7.36e^{-1}$} & \ttimesfm{} \\
 & Rank & $15.17$ & $14.83$ & $7.17$ & $7.83$ & $9.00$ & $14.33$ & $7.83$ & $17.00$ & $12.67$ & $6.67$ & $8.33$ & $17.50$ & $21.00$ & \underline{$4.83$} & $16.50$ & $7.50$ & $6.33$ & $7.00$ & $10.83$ & $15.50$ & \boldmath{$3.17$} & \tmoirailarge{} \\\hline
\multirow{1}{*}{Energy}& MAPE & 1.23 & 1.00 & 1.02 & 1.35 & 1.34 & 1.68 & 1.09 & 1.30 & 1.17 & 1.00 & 1.24 & 1.19 & $9.24e^{9}$ & $9.90e^{-1}$ & 1.15 & $9.11e^{-1}$ & $9.09e^{-1}$ & $9.12e^{-1}$ & \underline{$8.99e^{-1}$} & $9.87e^{-1}$ & \boldmath{$8.91e^{-1}$} & \tmoirailarge{} \\
 & CRPS & 1.72 & 1.00 & $8.63e^{-1}$ & 2.29 & $4.31e^{23}$ & 1.37 & $6.70e^{-1}$ & $8.82e^{-1}$ & $9.80e^{-1}$ & $6.52e^{-1}$ & $8.26e^{-1}$ & $9.16e^{-1}$ & $1.16e^{1}$ & $7.04e^{-1}$ & $8.25e^{-1}$ & $6.85e^{-1}$ & $6.61e^{-1}$ & $6.59e^{-1}$ & $6.97e^{-1}$ & \underline{$6.35e^{-1}$} & \boldmath{$6.26e^{-1}$} & \tmoirailarge{} \\
\bottomrule
\end{tabular}}%

\end{table}
\begin{table}
\Large

\caption{Results on \bench{} with all models aggregated by term length. The best results across each row are \textbf{bolded}, while second best results are \underline{underlined}.}
\label{tab:full_term_length_results}
\resizebox{\textwidth}{!}{%
\begin{tabular}{ll|lllll|llllllll|llllllll|l}
\toprule
Pred. Len. & Metric & \tnaive{} & \tsnaive{} & \tarima{} & \tthetaf{} & \tets{} & \tdeepar{} & \ttft{} & \ttide{} & \tnbeats{} & \tpatchtst{} & \titransformer{} & \tdlinear{} & \tcrossformer{} & \ttimesfm{} & \tvisionts{} & \tchronossmall{} & \tchronosbase{} & \tchronoslarge{} & \tmoiraismall{} & \tmoiraibase{} & \tmoirailarge{} & Best \\
\hline
\multirow{1}{*}{Long}& MAPE & 1.35 & 1.00 & 1.04 & 3.15 & 1.28 & 1.70 & 1.16 & 1.28 & 1.11 & \underline{$9.62e^{-1}$} & $9.96e^{-1}$ & 1.56 & $8.16e^{9}$ & 1.80 & 1.03 & 1.01 & $9.98e^{-1}$ & 1.00 & \boldmath{$9.19e^{-1}$} & 1.09 & 1.01 & \tmoiraismall{} \\
 & CRPS & 2.11 & 1.00 & $8.43e^{-1}$ & 2.06 & $6.57e^{23}$ & $9.48e^{-1}$ & \underline{$4.53e^{-1}$} & $5.25e^{-1}$ & $6.64e^{-1}$ & \boldmath{$4.37e^{-1}$} & $4.55e^{-1}$ & $6.73e^{-1}$ & $5.07e^{-1}$ & $6.54e^{-1}$ & $5.46e^{-1}$ & $6.46e^{-1}$ & $6.20e^{-1}$ & $6.21e^{-1}$ & $5.61e^{-1}$ & $5.01e^{-1}$ & $5.39e^{-1}$ & \tpatchtst{} \\
 & Rank & $19.62$ & $16.86$ & $15.10$ & $17.38$ & $18.67$ & $12.71$ & \underline{$4.90$} & $8.76$ & $11.95$ & \boldmath{$4.43$} & $5.19$ & $13.24$ & $8.29$ & $11.29$ & $9.57$ & $11.86$ & $10.71$ & $11.05$ & $7.24$ & $5.43$ & $6.76$ & \tpatchtst{} \\\hline
\multirow{1}{*}{Medium}& MAPE & 1.30 & 1.00 & $9.93e^{-1}$ & 1.87 & 1.59 & 1.07 & 1.26 & 1.19 & 1.15 & \boldmath{$9.03e^{-1}$} & 1.02 & 1.39 & $2.51e^{9}$ & 1.52 & 1.02 & 1.23 & 1.38 & 1.33 & \underline{$9.31e^{-1}$} & 1.11 & $9.60e^{-1}$ & \tpatchtst{} \\
 & CRPS & 2.08 & 1.00 & $8.63e^{-1}$ & 1.96 & $8.31e^{6}$ & $8.27e^{-1}$ & \underline{$5.29e^{-1}$} & $6.11e^{-1}$ & $7.60e^{-1}$ & \boldmath{$5.19e^{-1}$} & $5.29e^{-1}$ & $7.66e^{-1}$ & $6.61e^{-1}$ & $8.14e^{-1}$ & $6.69e^{-1}$ & $7.38e^{-1}$ & $7.77e^{-1}$ & $7.51e^{-1}$ & $6.61e^{-1}$ & $6.37e^{-1}$ & $6.25e^{-1}$ & \tpatchtst{} \\
\bottomrule
\end{tabular}}%
\end{table}

\begin{table}
\Large

\caption{Results on \bench{} with all models aggregated by frequency. The best results across each row are \textbf{bolded}, while second best results are \underline{underlined}.}
\label{tab:full_frequency_results}
\resizebox{\textwidth}{!}{%
\begin{tabular}{l|l|lllll|llllllll|llllllll|l}
\toprule
Freq. & Metric & \tnaive{} & \tsnaive{} & \tarima{} & \tthetaf{} & \tets{} & \tdeepar{} & \ttft{} & \ttide{} & \tnbeats{} & \tpatchtst{} & \titransformer{} & \tdlinear{} & \tcrossformer{} & \ttimesfm{} & \tvisionts{} & \tchronossmall{} & \tchronosbase{} & \tchronoslarge{} & \tmoiraismall{} & \tmoiraibase{} & \tmoirailarge{} & Best \\
\midrule
10S & CRPS & 1.54 & 1.00 & 1.00 & \boldmath{$4.50e^{-1}$} & 2.99 & 1.03 & 1.05 & $9.85e^{-1}$ & $8.21e^{-1}$ & $7.32e^{-1}$ & \underline{$6.91e^{-1}$} & 1.13 & $9.63e^{-1}$ & 1.82 & $9.34e^{-1}$ & 1.10 & 1.27 & 1.19 & 1.31 & 1.40 & 1.24 & \tthetaf{} \\
 & MAPE & 1.97 & 1.00 & 1.00 & \boldmath{$6.49e^{-1}$} & 1.77 & 1.83 & 3.24 & 1.67 & \underline{$9.97e^{-1}$} & 1.06 & 1.31 & 3.46 & $1.08e^{1}$ & 6.23 & 1.10 & 2.42 & 2.95 & 3.02 & 1.65 & 2.34 & 1.55 & \tthetaf{} \\
 & Rank & $16.33$ & $9.50$ & $8.50$ & \boldmath{$1.00$} & $19.50$ & $11.17$ & $7.67$ & $10.17$ & $6.67$ & $5.00$ & \underline{$2.50$} & $11.67$ & $11.50$ & $19.33$ & $10.17$ & $10.83$ & $12.33$ & $11.67$ & $14.83$ & $16.50$ & $14.17$ & \tthetaf{} \\
  \hline
5T & CRPS & 1.40 & 1.00 & 1.00 & 1.05 & 1.05 & $7.97e^{-1}$ & $5.57e^{-1}$ & $6.48e^{-1}$ & $7.26e^{-1}$ & $5.43e^{-1}$ & $5.43e^{-1}$ & $8.33e^{-1}$ & 1.04 & $7.30e^{-1}$ & $7.34e^{-1}$ & $7.52e^{-1}$ & $7.49e^{-1}$ & $7.55e^{-1}$ & \underline{$5.36e^{-1}$} & $5.41e^{-1}$ & \boldmath{$5.32e^{-1}$} & \tmoirailarge{} \\
 & MAPE & $8.44e^{-1}$ & 1.00 & 1.00 & $9.64e^{-1}$ & 1.01 & $6.72e^{-1}$ & $6.43e^{-1}$ & $7.95e^{-1}$ & $6.34e^{-1}$ & $5.87e^{-1}$ & $6.60e^{-1}$ & $9.70e^{-1}$ & 1.27 & $8.20e^{-1}$ & $8.76e^{-1}$ & $7.80e^{-1}$ & $7.63e^{-1}$ & $7.72e^{-1}$ & \boldmath{$5.30e^{-1}$} & $6.00e^{-1}$ & \underline{$5.52e^{-1}$} & \tmoiraismall{} \\
\bottomrule
\end{tabular}}%

\end{table}

\begin{table}
\Large

\caption{Results on \bench{} aggregated by number of variates. The best results across each row are \textbf{bolded}, while second best results are \underline{underlined}.}
\label{tab:full_variate_results}
\resizebox{\textwidth}{!}{%
\begin{tabular}{ll|lllll|llllllll|llllllll|l}
\toprule
Num. Var. & Metric & \tnaive{} & \tsnaive{} & \tarima{} & \tthetaf{} & \tets{} & \tdeepar{} & \ttft{} & \ttide{} & \tnbeats{} & \tpatchtst{} & \titransformer{} & \tdlinear{} & \tcrossformer{} & \ttimesfm{} & \tvisionts{} & \tchronossmall{} & \tchronosbase{} & \tchronoslarge{} & \tmoiraismall{} & \tmoiraibase{} & \tmoirailarge{} & Best \\
\midrule
\multirow{1}{*}{Multi\_v}& MAPE & 1.21 & 1.00 & $9.80e^{-1}$ & 2.41 & 1.32 & 1.54 & 1.40 & 1.44 & 1.43 & \boldmath{$8.75e^{-1}$} & 1.15 & 1.62 & 2.65 & 1.71 & 1.04 & 1.08 & 1.11 & 1.10 & \underline{$8.84e^{-1}$} & 1.09 & $9.32e^{-1}$ & \tpatchtst{} \\
 & CRPS & 1.40 & 1.00 & $8.71e^{-1}$ & 1.19 & $2.82e^{7}$ & 1.18 & \underline{$6.09e^{-1}$} & $8.20e^{-1}$ & $7.35e^{-1}$ & \boldmath{$5.27e^{-1}$} & $6.22e^{-1}$ & $7.82e^{-1}$ & $9.19e^{-1}$ & $7.50e^{-1}$ & $7.01e^{-1}$ & $6.65e^{-1}$ & $6.83e^{-1}$ & $6.73e^{-1}$ & $6.26e^{-1}$ & $6.28e^{-1}$ & $6.27e^{-1}$ & \tpatchtst{} \\
 & Rank & $17.70$ & $16.51$ & $13.91$ & $14.98$ & $17.02$ & $14.00$ & $6.91$ & $11.40$ & $12.58$ & \boldmath{$5.02$} & \underline{$5.19$} & $13.49$ & $12.91$ & $10.07$ & $11.53$ & $9.51$ & $9.44$ & $9.56$ & $6.33$ & $6.28$ & $6.67$ & \tpatchtst{} \\\hline
\multirow{1}{*}{Uni\_v}& MAPE & 1.19 & 1.00 & $9.45e^{-1}$ & 1.18 & 1.12 & $9.49e^{-1}$ & $8.92e^{-1}$ & $9.60e^{-1}$ & \boldmath{$7.96e^{-1}$} & $8.47e^{-1}$ & $8.53e^{-1}$ & 1.02 & $5.47e^{9}$ & $8.72e^{-1}$ & $9.34e^{-1}$ & $8.11e^{-1}$ & $8.06e^{-1}$ & \underline{$7.99e^{-1}$} & $8.80e^{-1}$ & $9.47e^{-1}$ & $8.10e^{-1}$ & \tnbeats{} \\
 & CRPS & 1.74 & 1.00 & $7.62e^{-1}$ & 1.59 & $2.55e^{23}$ & $8.02e^{-1}$ & \underline{$5.85e^{-1}$} & $7.23e^{-1}$ & $8.03e^{-1}$ & $5.92e^{-1}$ & $6.28e^{-1}$ & $8.29e^{-1}$ & $2.88e^{1}$ & $6.30e^{-1}$ & $7.49e^{-1}$ & $6.18e^{-1}$ & $6.00e^{-1}$ & $5.95e^{-1}$ & $6.55e^{-1}$ & $6.63e^{-1}$ & \boldmath{$5.66e^{-1}$} & \tmoirailarge{} \\
\bottomrule
\end{tabular}}%
\end{table}

\begin{table}
\Large

\caption{Results on \bench{} with all models aggregated by all datasets. The best results across each row are \textbf{bolded}, while second best results are \underline{underlined}.}
\label{tab:full_results}
\resizebox{\textwidth}{!}{%
\begin{tabular}{l|lllll|llllllll|llllllll|l}
\toprule
Metric & \tnaive{} & \tsnaive{} & \tarima{} & \tthetaf{} & \tets{} & \tdeepar{} & \ttft{} & \ttide{} & \tnbeats{} & \tpatchtst{} & \titransformer{} & \tdlinear{} & \tcrossformer{} & \ttimesfm{} & \tvisionts{} & \tchronossmall{} & \tchronosbase{} & \tchronoslarge{} & \tmoiraismall{} & \tmoiraibase{} & \tmoirailarge{} & Best \\

MAPE & 1.20 & 1.00 & $9.61e^{-1}$ & 1.73 & 1.20 & 1.21 & 1.12 & 1.17 & 1.08 & \boldmath{$8.60e^{-1}$} & $9.85e^{-1}$ & 1.28 & $3.05e^{9}$ & 1.25 & $9.82e^{-1}$ & $9.28e^{-1}$ & $9.40e^{-1}$ & $9.30e^{-1}$ & $8.82e^{-1}$ & 1.01 & \underline{$8.64e^{-1}$} & \tpatchtst{} \\
CRPS & 1.59 & 1.00 & $8.11e^{-1}$ & 1.41 & $1.42e^{23}$ & $9.69e^{-1}$ & $5.96e^{-1}$ & $7.66e^{-1}$ & $7.73e^{-1}$ & \boldmath{$5.63e^{-1}$} & $6.26e^{-1}$ & $8.08e^{-1}$ & $1.65e^{1}$ & $6.83e^{-1}$ & $7.28e^{-1}$ & $6.38e^{-1}$ & $6.37e^{-1}$ & $6.29e^{-1}$ & $6.42e^{-1}$ & $6.48e^{-1}$ & \underline{$5.93e^{-1}$} & \tpatchtst{} \\
Rank & $18.19$ & $16.64$ & $13.15$ & $15.21$ & $15.82$ & $11.77$ & $6.76$ & $11.57$ & $13.26$ & \boldmath{$5.72$} & $6.37$ & $14.45$ & $13.62$ & $8.55$ & $12.75$ & $9.16$ & $8.32$ & $8.31$ & $7.81$ & $7.57$ & \underline{$5.99$} & \tpatchtst{} \\
\bottomrule
\end{tabular}}%

\end{table}

\begin{table}
\centering
\Huge

\caption{Results on all dataset configs for \bench{} | Table 1/3. The best results across each row are \textbf{bolded}, while second best results are \underline{underlined}.}
\label{tab:all_results}
\resizebox{\textwidth}{!}{%
\begin{tabular}{ll|lllll|llllllll|llllllll|l}
\toprule
Dataset,term,frequency & Metric & \tnaive{} & \tsnaive{} & \tarima{} & \tthetaf{} & \tets{} & \tdeepar{} & \ttft{} & \ttide{} & \tnbeats{} & \tpatchtst{} & \titransformer{} & \tdlinear{} & \tcrossformer{} & \ttimesfm{} & \tvisionts{} & \tchronossmall{} & \tchronosbase{} & \tchronoslarge{} & \tmoiraismall{} & \tmoiraibase{} & \tmoirailarge{} & Best \\

bitbrains\_fast\_storage, long, 5T & MAPE & $2.79e^{-1}$ & 1.00 & 1.00 & $5.79e^{-1}$ & 1.00 & \boldmath{$1.90e^{-1}$} & 1.04 & $5.12e^{-1}$ & $2.45e^{-1}$ & $4.10e^{-1}$ & $5.79e^{-1}$ & 1.34 & 1.14 & $6.36e^{-1}$ & $9.23e^{-1}$ & $2.83e^{-1}$ & \underline{$2.29e^{-1}$} & $2.49e^{-1}$ & $4.15e^{-1}$ & $4.97e^{-1}$ & $5.37e^{-1}$ & \tdeepar{} \\
bitbrains\_fast\_storage, long, 5T & CRPS & 1.64 & 1.00 & 1.00 & 1.05 & 1.00 & $7.83e^{-1}$ & $5.69e^{-1}$ & $5.47e^{-1}$ & $6.13e^{-1}$ & $5.19e^{-1}$ & $5.35e^{-1}$ & 1.03 & $8.22e^{-1}$ & $6.25e^{-1}$ & $7.50e^{-1}$ & $5.50e^{-1}$ & $5.51e^{-1}$ & $5.60e^{-1}$ & \underline{$5.11e^{-1}$} & \boldmath{$4.92e^{-1}$} & $5.17e^{-1}$ & \tmoiraibase{} \\
bitbrains\_fast\_storage, long, 5T & Rank & $21.00$ & $18.00$ & $16.00$ & $20.00$ & $17.00$ & $14.00$ & $10.00$ & $6.00$ & $11.00$ & $4.00$ & $5.00$ & $19.00$ & $15.00$ & $12.00$ & $13.00$ & $7.00$ & $8.00$ & $9.00$ & \underline{$2.00$} & \boldmath{$1.00$} & $3.00$ & \tmoiraibase{} \\
bitbrains\_fast\_storage, medium, 5T & MAPE & $7.14e^{-1}$ & 1.00 & 1.00 & $8.47e^{-1}$ & 1.00 & \boldmath{$1.92e^{-1}$} & $3.85e^{-1}$ & $7.07e^{-1}$ & \underline{$2.53e^{-1}$} & $5.48e^{-1}$ & $6.93e^{-1}$ & $8.27e^{-1}$ & $9.74e^{-1}$ & $6.52e^{-1}$ & $9.43e^{-1}$ & $6.50e^{-1}$ & $5.74e^{-1}$ & $5.75e^{-1}$ & $5.77e^{-1}$ & $6.53e^{-1}$ & $5.79e^{-1}$ & \tdeepar{} \\
\bottomrule
\end{tabular}}
\end{table}

\begin{table}
\centering
\Huge

\caption{Results on all dataset configs for \bench{} | Table 2/3. The best results across each row are \textbf{bolded}, while second best results are \underline{underlined}.}
\label{tab:all_results2}
\resizebox{\textwidth}{!}{%
\begin{tabular}{ll|lllll|llllllll|llllllll|l}
\toprule
Dataset,term,frequency & Metric & \tnaive{} & \tsnaive{} & \tarima{} & \tthetaf{} & \tets{} & \tdeepar{} & \ttft{} & \ttide{} & \tnbeats{} & \tpatchtst{} & \titransformer{} & \tdlinear{} & \tcrossformer{} & \ttimesfm{} & \tvisionts{} & \tchronossmall{} & \tchronosbase{} & \tchronoslarge{} & \tmoiraismall{} & \tmoiraibase{} & \tmoirailarge{} & Best \\

ett1, long, 15T & MAPE & 1.19 & 1.00 & 1.00 & 1.20 & 1.33 & 4.72 & $8.37e^{-1}$ & $8.10e^{-1}$ & \underline{$7.82e^{-1}$} & $8.37e^{-1}$ & $8.47e^{-1}$ & $8.20e^{-1}$ & $8.04e^{-1}$ & $8.12e^{-1}$ & $8.29e^{-1}$ & $8.88e^{-1}$ & $8.60e^{-1}$ & $8.72e^{-1}$ & $8.78e^{-1}$ & $8.50e^{-1}$ & \boldmath{$7.72e^{-1}$} & \tmoirailarge{} \\
ett1, long, 15T & CRPS & 1.81 & 1.00 & 1.00 & 3.51 & 2.65 & 5.61 & $7.07e^{-1}$ & $7.27e^{-1}$ & 1.09 & \underline{$6.24e^{-1}$} & \boldmath{$6.19e^{-1}$} & $8.66e^{-1}$ & $8.64e^{-1}$ & $9.04e^{-1}$ & $8.08e^{-1}$ & 1.14 & 1.01 & $9.27e^{-1}$ & $7.20e^{-1}$ & $6.59e^{-1}$ & $7.58e^{-1}$ & \titransformer{} \\
ett1, long, 15T & Rank & $18.00$ & $14.00$ & $13.00$ & $20.00$ & $19.00$ & $21.00$ & $4.00$ & $6.00$ & $16.00$ & \underline{$2.00$} & \boldmath{$1.00$} & $10.00$ & $9.00$ & $11.00$ & $8.00$ & $17.00$ & $15.00$ & $12.00$ & $5.00$ & $3.00$ & $7.00$ & \titransformer{} \\
ett1, long, H & MAPE & 2.13 & 1.00 & $8.26e^{-1}$ & 1.01 & 2.13 & 1.50 & 1.00 & 1.48 & \underline{$7.66e^{-1}$} & 1.69 & 1.17 & 1.22 & 1.02 & \boldmath{$7.25e^{-1}$} & 1.45 & $9.58e^{-1}$ & 1.11 & 1.17 & $9.74e^{-1}$ & 1.77 & 1.19 & \ttimesfm{} \\
ett1, long, H & CRPS & 1.60 & 1.00 & $6.98e^{-1}$ & 3.15 & $7.27e^{2}$ & $7.61e^{-1}$ & $5.08e^{-1}$ & $5.08e^{-1}$ & $9.20e^{-1}$ & $4.82e^{-1}$ & \underline{$4.82e^{-1}$} & $5.89e^{-1}$ & $9.25e^{-1}$ & $5.15e^{-1}$ & $5.91e^{-1}$ & $5.84e^{-1}$ & $5.68e^{-1}$ & $5.68e^{-1}$ & \boldmath{$4.63e^{-1}$} & $4.98e^{-1}$ & $5.75e^{-1}$ & \tmoiraismall{} \\
\bottomrule
\end{tabular}}
\end{table}

\begin{table}
\centering
\Huge

\caption{Results on all dataset configs for \bench{} | Table 3/3. The best results across each row are \textbf{bolded}, while second best results are \underline{underlined}.}
\label{tab:all_results3}
\resizebox{\textwidth}{!}{%
\begin{tabular}{ll|lllll|llllllll|llllllll|l}
\toprule
Dataset,term,frequency & Metric & \tnaive{} & \tsnaive{} & \tarima{} & \tthetaf{} & \tets{} & \tdeepar{} & \ttft{} & \ttide{} & \tnbeats{} & \tpatchtst{} & \titransformer{} & \tdlinear{} & \tcrossformer{} & \ttimesfm{} & \tvisionts{} & \tchronossmall{} & \tchronosbase{} & \tchronoslarge{} & \tmoiraismall{} & \tmoiraibase{} & \tmoirailarge{} & Best \\

loop\_seattle, short, 5T & MAPE & 1.06 & 1.00 & 1.00 & $9.00e^{-1}$ & 1.02 & $9.85e^{-1}$ & $9.38e^{-1}$ & 1.09 & $9.62e^{-1}$ & 1.04 & 1.07 & 1.12 & 2.21 & 1.19 & 1.15 & $9.85e^{-1}$ & $9.77e^{-1}$ & $9.77e^{-1}$ & $8.69e^{-1}$ & \underline{$8.38e^{-1}$} & \boldmath{$8.31e^{-1}$} & \tmoirailarge{} \\
loop\_seattle, short, 5T & CRPS & 1.44 & 1.00 & 1.00 & 1.01 & 1.03 & $8.91e^{-1}$ & $8.01e^{-1}$ & $9.01e^{-1}$ & $9.89e^{-1}$ & $8.12e^{-1}$ & $8.12e^{-1}$ & 1.22 & 5.64 & $9.29e^{-1}$ & 1.08 & $8.66e^{-1}$ & $8.61e^{-1}$ & $8.64e^{-1}$ & $6.52e^{-1}$ & \boldmath{$6.44e^{-1}$} & \underline{$6.51e^{-1}$} & \tmoiraibase{} \\
loop\_seattle, short, 5T & Rank & $20.00$ & $15.00$ & $14.00$ & $16.00$ & $17.00$ & $10.00$ & $4.00$ & $11.00$ & $13.00$ & $6.00$ & $5.00$ & $19.00$ & $21.00$ & $12.00$ & $18.00$ & $9.00$ & $7.00$ & $8.00$ & $3.00$ & \boldmath{$1.00$} & \underline{$2.00$} & \tmoiraibase{} \\
loop\_seattle, short, D & MAPE & 1.00 & 1.00 & $8.50e^{-1}$ & $8.01e^{-1}$ & $7.07e^{-1}$ & $5.58e^{-1}$ & $5.32e^{-1}$ & $5.08e^{-1}$ & $4.96e^{-1}$ & $5.16e^{-1}$ & $5.06e^{-1}$ & $4.99e^{-1}$ & $8.21e^{-1}$ & \boldmath{$4.88e^{-1}$} & $5.29e^{-1}$ & $5.56e^{-1}$ & $5.04e^{-1}$ & $5.04e^{-1}$ & $5.36e^{-1}$ & $5.17e^{-1}$ & \underline{$4.94e^{-1}$} & \ttimesfm{} \\
\bottomrule
\end{tabular}}
\end{table}

\begin{table}
\tiny
\caption{\moirai{} vs \omoirai{} results on datasets that LOTSA collection and our \bench{} has in common.}
\label{tab:leak_results}
\resizebox{\textwidth}{!}{%
\begin{tabular}{c|c|cc|cc|cc}
\hline
\multirow{2}{*}{Dataset} & \multirow{2}{*}{Model} & \multicolumn{2}{c|}{Short} & \multicolumn{2}{c|}{Medium} & \multicolumn{2}{c}{Long} \\ \cline{3-8} 
                      &                                   & MAPE & CRPS & MAPE & CRPS & MAPE & CRPS \\ \hline
hierarchical\_sales, D & \tomoiraibase{}  & 0.51 & 0.24 & NA   & NA   & NA   & NA   \\ \hline
hierarchical\_sales, D & \tmoiraibase{}   & 0.49 & 0.25 & NA   & NA   & NA   & NA   \\ \hline
hierarchical\_sales, D & \tomoirailarge{} & 0.53 & 0.25 & NA   & NA   & NA   & NA   \\ \hline
hierarchical\_sales, D & \tmoirailarge{}  & 0.52 & 0.24 & NA   & NA   & NA   & NA   \\ \hline
hierarchical\_sale, D & \tomoiraismall{} & 0.50 & 0.25 & NA   & NA   & NA   & NA   \\ \hline
hierarchical\_sales, D & \tmoiraismall{}  & 0.51 & 0.25 & NA   & NA   & NA   & NA   \\
\hline
\end{tabular}%
}
\end{table}

\end{document}